\documentclass{article}
\PassOptionsToPackage{round}{natbib}
\usepackage[eandd, final]{neurips_2026}

\usepackage[utf8]{inputenc} 
\usepackage[T1]{fontenc}    
\usepackage[hidelinks]{hyperref}       
\usepackage{url}            
\usepackage{booktabs}       
\usepackage{amsfonts}       
\usepackage{nicefrac}       
\usepackage{microtype}      
\usepackage{xcolor}         

\usepackage{booktabs}
\usepackage{multirow}
\usepackage{enumitem}
\usepackage{graphicx}
\usepackage{subcaption}
\usepackage{amsmath}
\usepackage{mdframed}
\usepackage{ulem}

\usepackage[most]{tcolorbox}

\tcbset{frametitle/.style={title={#1}}}

\newtcolorbox{custombox}[1][]{
    breakable,
    enhanced,
    parbox=true,
    before upper={\setlength{\parskip}{0.5\baselineskip}},
    colback=white,
    colframe=black,
    boxrule=0.5pt,
    arc=2pt,
    #1
}

\title{
Reporting Practice Matters: The Impact of Reference Choice on Chest X-ray Report Evaluation
}

\author{%
  Daniel P. Jeong\textsuperscript{1}, 
  Charles Q. Li\textsuperscript{2}, 
  Hossein Hosseiny\textsuperscript{3}, 
  Nitya M. Bhalla\textsuperscript{2}, \\
  \textbf{%
    Fatma Uyar Morency\textsuperscript{3}, 
    Pradeep Ravikumar\textsuperscript{1}, 
    Zachary C. Lipton\textsuperscript{1},
    Michael Oberst\textsuperscript{4}%
  } \\
  \textsuperscript{1} Machine Learning Department, Carnegie Mellon University \\
  \textsuperscript{2} Department of Radiology and Imaging Sciences, Allegheny Health Network \\
  \textsuperscript{3} Data Science R\&D, Highmark Health Enterprise Data \& Analytics \\
  \textsuperscript{4} Department of Computer Science, Johns Hopkins University \\
}

\begin{document}


\maketitle


\begin{abstract}
Radiologists follow heterogeneous \textit{reporting practices}. Two radiologists examining the same image and identifying the same \textit{clinical findings} might nevertheless compose superficially distinct reports, varying in terminology, shorthand, formatting, and level of detail. These variations in reporting norms represent an under-appreciated obstacle in efforts to evaluate AI-based radiology report generation (RRG) models, where machine-generated reports are typically assessed based on their concordance with human-generated references. In this paper, we quantify the sensitivity of established evaluation metrics to variations in reporting practices, revealing impacts large enough to alter the rankings of models. We introduce a radiologist-informed taxonomy of variations in radiology reporting practice and a method (\textsc{ReRef}) that rewrites reference reports along the axes of our taxonomy while preserving clinical interpretation. For instance, when comparing the performance of nine RRG models on MIMIC-CXR using RadCliQ-v1, condensing the discussion of normal findings in the reference reports causes Libra to drop from first to second place while CheXOne rises from third to first. Our results suggest that many current metrics fail to decouple clinical interpretation from conformity to reporting practices and that choosing the ``right'' references that accurately reflect the desired reporting practices can be important in practice. To support future research, we release \textsc{MIMIC-CXR-Ext-ReRef}, a radiologist-validated dataset of 120 (original, alternative) reference report pairs derived from MIMIC-CXR.
\end{abstract}

\section{Introduction}
\label{sec:intro}
Automated generation of radiology reports, the primary medium through which radiologists document their interpretation of medical images (e.g., chest X-ray (CXR), computed tomography (CT) scans), has the potential to improve the efficiency of radiologist workflows, helping reduce physician burnout and improve patient care. 
To realize this potential, several works propose to train vision-language models (VLMs) that are capable of generating a text report summarizing the key clinical findings (i.e., the Findings and/or Impression sections of a radiology report) when provided with the radiological images \citep{unirg-cxr,medversa,maira-2}. 
While some studies suggest that they are starting to show promise in assisting radiologists under controlled setups \citep{flamingo-cxr,northwestern-study}, 
recent benchmarks indicate that the current crop of models are not yet reliable or accurate enough for deployment in real-world settings \citep{automated-cxr-unsolved,rexrank}.

Given the free-form nature of radiology reports, a major challenge to model development lies in designing evaluation methods that reliably characterize the clinical accuracy and utility of model-generated reports. 
As manual review by a radiologist is not scalable, a common approach is to employ an \textit{automated} evaluation metric (e.g., GREEN \citep{green}, RadCliQ-v1 \citep{radcliq}, RadGraph-F1 \citep{radgraph-f1}) that compares the model-generated report to a \textit{reference} report, written independently by a radiologist, and outputs a numerical score (e.g., between 0 and 1). 
State-of-the-art metrics are claimed to correlate well with radiologist judgment of report quality, e.g., as measured against radiologist annotations in datasets like ReXVal \citep{radcliq}.

\begin{figure*}[!t]
    \centering
    \begin{tabular}{c@{\hskip 5pt}c@{}}
        \begin{subfigure}{0.545\linewidth} 
            \includegraphics[width=\linewidth]{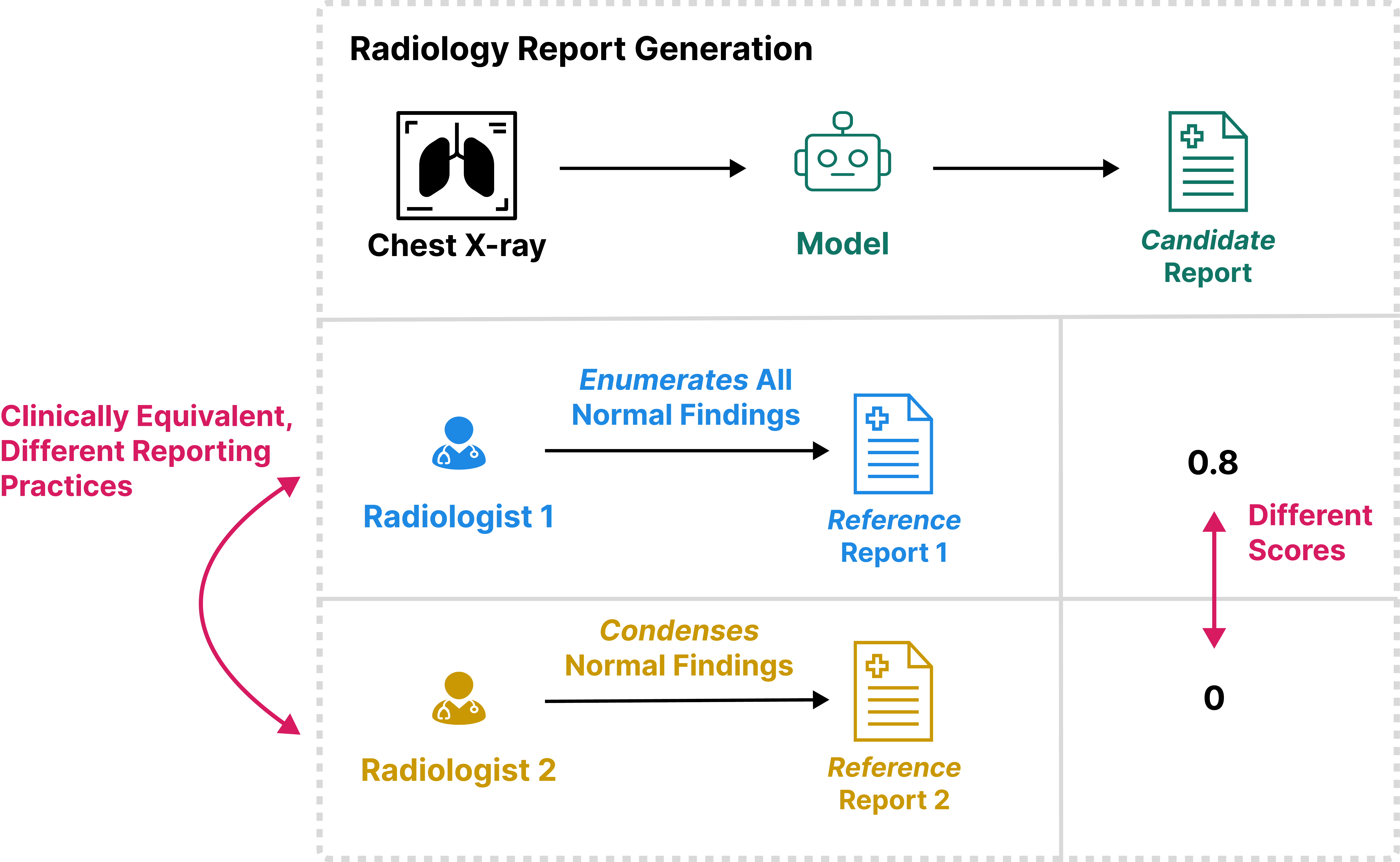}
        \end{subfigure} &
        \begin{subfigure}{0.425\linewidth} 
            \includegraphics[width=\linewidth]{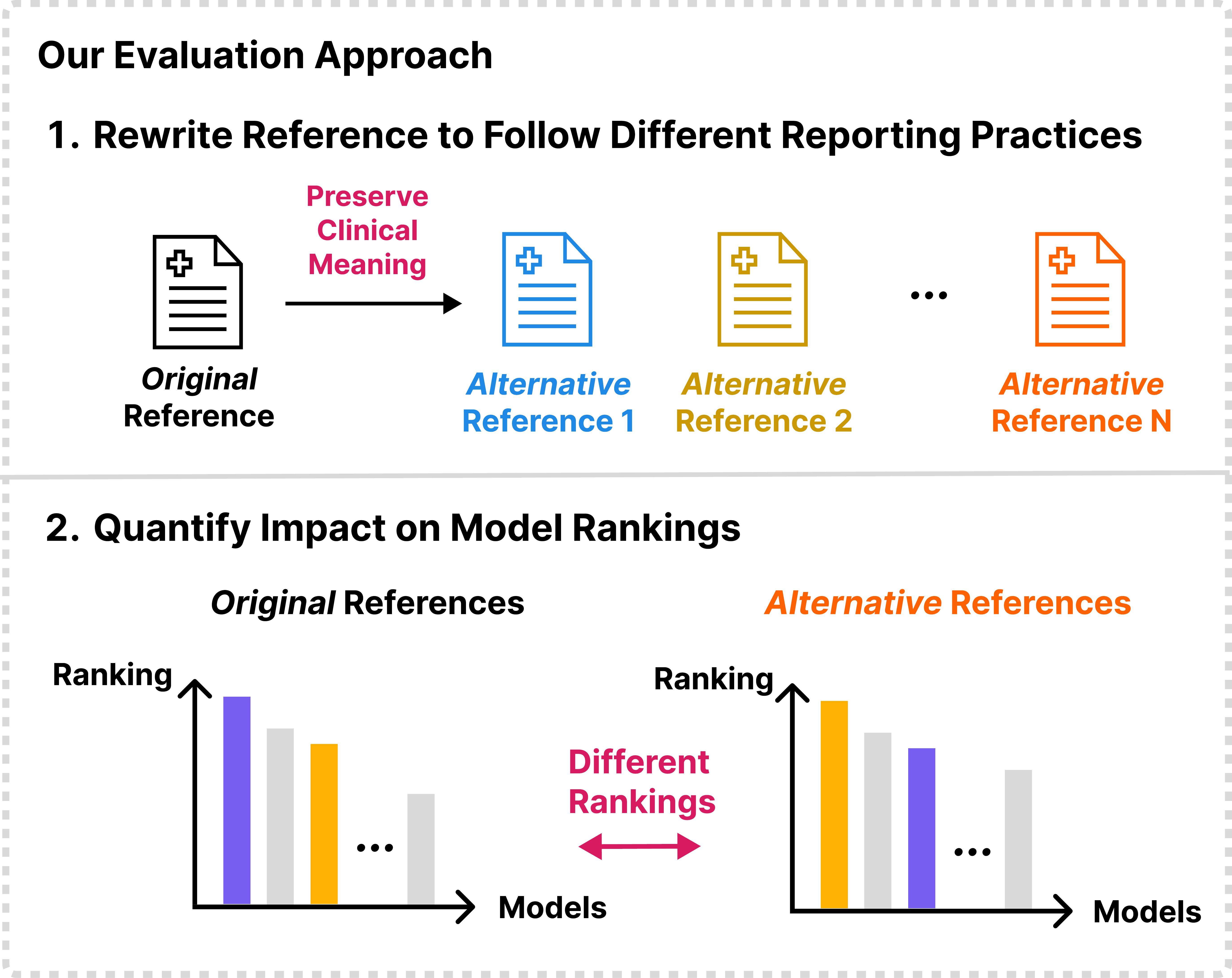}
        \end{subfigure}
    \end{tabular}
    \caption{Motivation and overview. 
    (Left) Automated evaluation metrics for radiology report generation (RRG) models can be sensitive to the reporting practices of the reference: when individually compared against two references that communicate \textit{identical clinical findings} but follow distinct reporting practices, a model-generated report may score higher against one vs.~the other by better conforming to the reporting practices of the former (e.g., enumerating all normal findings). 
    (Right) To study how these variations impact the conclusions of RRG model evaluation, 
    we propose a method (\textsc{ReRef}; Section~\ref{sec:reref}) that rewrites a reference report to follow alternative reporting practices while preserving clinical interpretation, and quantify the impact of each \textit{perturbation} on the model rankings.
    }
    \label{fig:main-figure}
    \vspace{-10pt}
\end{figure*}

However, radiologists follow highly heterogeneous reporting practices, often communicating interpretations with similar clinical impact in varying language and levels of detail \citep{srrg-nightmare,srrg-fuel,structured-rrg}, 
which raises concerns about whether current evaluation approaches are sensitive to these differences. 
For instance, 
a model-generated report that enumerates normal findings (e.g., ``The lungs are clear'', ``There is no pleural effusion'') may receive a higher score  
when evaluated against a similarly enumerated reference (e.g., ``Lungs are clear'', ``No pleural effusion or pneumothorax evident''), 
than against a more abstract and lexically different reference report (``No acute cardiothoracic abnormality'') (Figure~\ref{fig:main-figure}). 

Motivated by such observations, we investigate how variations in reporting practice can impact the conclusions about radiology report generation (RRG) model performance, using nine open-source models and four datasets for CXRs. 
We construct a taxonomy that characterizes the salient ways in which radiology reporting practice differs (e.g., report structure, level of detail, terminology preferences) (Section~\ref{sec:taxonomy}). 
We then develop \textsc{ReRef}, an expert-validated method that uses a large language model (LLM) to rewrite reference reports along the axes of our taxonomy 
while preserving the overall clinical interpretation (Sections~\ref{sec:reref}--\ref{sec:annotation}, \ref{sec:results-reref}). 
We use \textsc{ReRef} to \textit{perturb} the reference reports of all datasets 
to study how reporting practice variations impact model performance 
(Section~\ref{sec:experiments}).

Overall, we find that many existing metrics are sensitive to such variations and 
fail to decouple clinical interpretation 
from conformity to reporting practices (Section~\ref{sec:results-metric-sensitivity}), 
large enough to alter the rankings of models 
(Section~\ref{sec:results-nonuniform-impact}). 
For example,
when comparing all models on MIMIC-CXR \citep{mimic-cxr} with RadCliQ-v1, 
condensing the reference reports to focus on the key abnormal findings causes 
Libra \citep{libra} to drop from first to second place in model ranking, 
while CheXOne \citep{chexone} rises from 
third to first place. 
Our findings suggest that, in practice, selecting references that best-reflect the desired reporting practices for the setting of interest can be an important design decision for model evaluation.
We also release \textsc{MIMIC-CXR-Ext-ReRef}, a radiologist-validated dataset of 120 (original, alternative) report pairs derived from MIMIC-CXR. 

Our main contributions can be summarized as follows:
\begin{enumerate}[itemsep=0.1ex,topsep=0.1ex]
    \item We present a taxonomy that systematically captures the variations in radiology reporting practice, designed under the guidance of a board-certified radiologist (Section~\ref{sec:taxonomy}).
    \item We develop \textsc{ReRef}, a method for rewriting a given reference report along each axis of our taxonomy (Section~\ref{sec:reref}), and validate it via radiologist annotations (Sections~\ref{sec:annotation} and \ref{sec:results-reref}).
    \item We propose an evaluation framework based on \textsc{ReRef} to study how variations in radiology reporting practice impact the conclusions about RRG model performance (Section~\ref{sec:experiments}).
    \item We find that many current metrics for RRG model evaluation fail to decouple clinical interpretation from conformity to reporting practices of the reference (Sections~\ref{sec:results-metric-sensitivity}--\ref{sec:results-nonuniform-impact}).
    \item We release \textsc{MIMIC-CXR-Ext-ReRef}\footnote{\href{https://physionet.org/content/mimic-cxr-ext-reref/1.0.0/}{physionet.org/content/mimic-cxr-ext-reref/1.0.0/}}, a radiologist-validated dataset of 120 (original, alternative) reference report pairs derived from MIMIC-CXR to support future research.
\end{enumerate}

To ensure the reproducibility of our results, we open-source the source code used for \textsc{ReRef} and for all of our evaluations described below via our GitHub repository\footnote{\href{https://github.com/taekb/rrg-reref}{github.com/taekb/rrg-reref}}.

\section{Categorizing and Simulating Reporting Practice Variation}
\label{sec:reporting-practice-variation}
To study how reporting practice variation impacts the evaluation of RRG models, we first construct a taxonomy of such variation under the guidance of a board-certified radiologist (Section~\ref{sec:taxonomy}). 
We then introduce \textsc{ReRef}, an LLM-based method that rewrites a reference report along a chosen axis of our taxonomy (Section~\ref{sec:reref}). 
Finally, we conduct an annotation study to validate that \textsc{ReRef} produces alternatives that are \textit{realistic} (i.e., plausibly written by a radiologist) and \textit{clinically equivalent} to the original reference (i.e., overall diagnosis and implied management remain unchanged) 
(Section~\ref{sec:annotation}).

\vspace{-5pt}
\subsection{Taxonomy of Reporting Practice Variation}
\label{sec:taxonomy}
\vspace{-5pt}

We categorize the variations in radiology reporting practice along four \textit{dimensions}---\textit{Organization}, \textit{Completeness}, \textit{Granularity}, and \textit{Language}---where each comprises one or more specific \textit{axes} along which radiology reports often differ without substantial change in clinical interpretation. 
For example, under \textit{Granularity}, we define the \textit{Anatomical Granularity} and \textit{Quantitative Granularity} axes, which each capture a specific way in which radiologists can control the level of detail in their reports. 

\vspace{-7pt}
\paragraph{Organization.} \textit{Organization} captures how report content is arranged at the report level and comprises two axes: \textit{Structure} and \textit{Section Assignment}.
\textit{Structure} contrasts highly templated formats---e.g., use of explicit anatomical headings, bullet points, numbered lists, subheadings for groups of findings (\textit{Structure (Structured)})---with continuous free-text prose (\textit{Structure (Free-Text)}). 
\textit{Section Assignment} captures the variations in how radiologists allocate content to different sections of a radiology report.
For example, comparisons to prior studies and differential diagnoses may be discussed in the Findings, in the Impression, or in both, depending on the radiologist's preference.

\vspace{-7pt}
\paragraph{Completeness.} \textit{Completeness} captures how exhaustively a report describes the findings visible in a study. 
We focus on a specific instantiation of this dimension, \textit{Findings Scope}, which contrasts an exhaustive style that enumerates all normal and abnormal findings (\textit{Findings Scope (Exhaustive)}), against a minimal style that selectively omits normal findings and reports only a focused list of abnormalities most likely to be clinically relevant (\textit{Findings Scope (Minimal)}). 

\vspace{-7pt}
\paragraph{Granularity.} \textit{Granularity} captures the level of detail at which findings are reported. 
\textit{Anatomical Granularity} contrasts reports that enumerate findings for each anatomical substructure (e.g., ``pulmonary nodule in the right upper lobe anterior segment'') (\textit{Anatomical Granularity (High)}) with those that aggregate findings at the organ/system level (e.g., ``right lung pulmonary nodule'') (\textit{Anatomical Granularity (Low)}). 
\textit{Quantitative Granularity} contrasts reports that include explicit numerical measurements (e.g., ``cardiothoracic ratio of 0.55'') (\textit{Quantitative Granularity (High)}), with those that provide qualitative descriptions 
(e.g., ``mild cardiomegaly'') (\textit{Quantitative Granularity (Low)}).

\vspace{-7pt}
\paragraph{Language.} \textit{Language} captures the linguistic variation in how radiologists report their interpretations. 
\textit{Terminology} captures substitution of near-synonymous phrasings (e.g., ``cardiomegaly'' vs.~``enlarged cardiac silhouette''), reflecting differences in training background, subspecialty, or institutional convention. 
\textit{Hedging} captures how diagnostic uncertainty is expressed at an approximately fixed underlying confidence level (e.g., ``worrisome for'' vs.~``concerning for''; ``consistent with'' vs.~``likely represents''). \textit{Negation} contrasts explicit negation of specific findings (e.g., ``no pneumothorax, no pleural effusion'') (\textit{Negation (Explicit)}), with implicit phrasings that convey the same information through positive statements about normality (e.g., ``pleura appear normal'') (\textit{Negation (Implicit)}).

\subsection{\textsc{ReRef}: Rewriting a Reference Report to Follow Different Reporting Practices}
\label{sec:reref}
\vspace{-5pt}

\begin{figure*}[t!]
\centering
\begin{tabular}{c@{}c}
    \begin{subfigure}{0.03\linewidth}
        \makebox[\linewidth]{\raisebox{65pt}{{(a)}}} 
    \end{subfigure} &
    \begin{subfigure}{0.85\linewidth} 
        \includegraphics[width=0.9\linewidth]{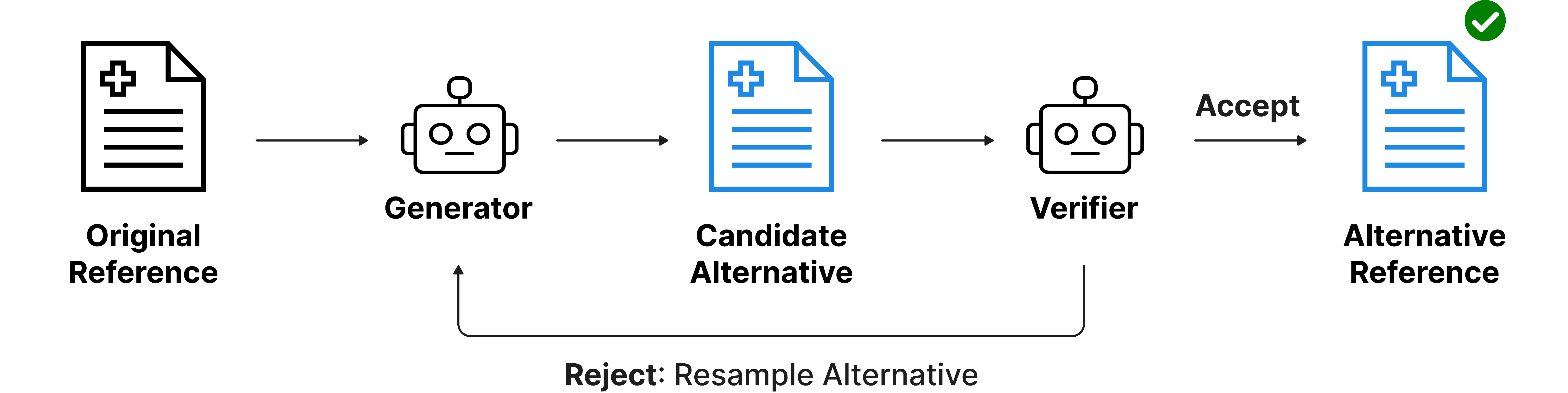}
    \end{subfigure} \\
    \begin{subfigure}{0.03\linewidth}
        \makebox[\linewidth]{\raisebox{55pt}{{(b)}}} 
    \end{subfigure} &
    \begin{subfigure}{0.85\linewidth} 
        \includegraphics[width=0.9\linewidth]{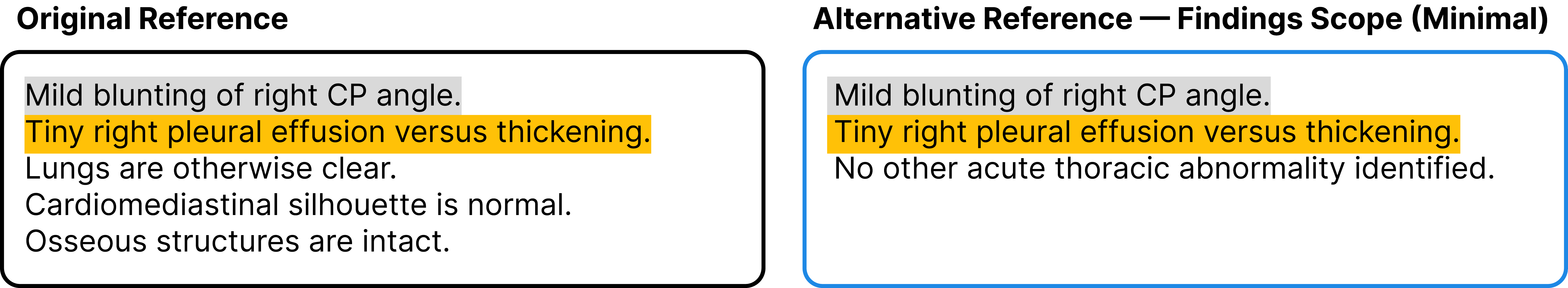}
    \end{subfigure}
\end{tabular}
\caption{Overview of \textsc{ReRef} (Section~\ref{sec:reref})---our method for rewriting a reference report to follow an alternative reporting practice aligned with an axis of our taxonomy (Section~\ref{sec:taxonomy}). (a) Given a reference report and the target axis-aligned variation (e.g., \textit{Findings Scope (Minimal)}), the \textit{Generator} produces an alternative reference. The \textit{Verifier} checks whether (i) the generated alternative report is clinically equivalent to the original report (e.g., no hallucinated findings) and (ii) the intended variation is accurately simulated. The generation--verification loop repeats until either a valid alternative is generated or the max number of generation attempts is reached. 
(b) Example output for \textit{Findings Scope (Minimal)}, which condenses the normal findings and focuses discussion on the key abnormalities. The highlighted findings are the abnormalities that stay preserved.}
\label{fig:reref}
\vspace{-5pt}
\end{figure*}

Based on the proposed taxonomy in Section~\ref{sec:taxonomy}, we develop our \textsc{ReRef} method, which takes as input a reference report (all sections) and an \textit{axis-aligned} target variation (e.g., \textit{Terminology}, \textit{Structure}) and outputs an \textit{alternative} reference rewritten to follow the specified reporting practices (Figure~\ref{fig:reref}). 
We implement \textsc{ReRef} by instantiating a \textit{Generator} and a \textit{Verifier}, which are each in charge of drafting the axis-aligned alternative reference, and validating if the generated report (i) accurately simulates the target variation, and (ii) is clinically equivalent to the original reference---i.e., the overall diagnosis of the patient and implied management plans do not change.

Notably, the \textit{Verifier} also checks whether the target variation is \textit{feasible} for the input report, as certain variations are impossible to simulate without hallucinating new clinical information (e.g., \textit{Quantitative Granularity (High)} would require adding new numerical measurements which are not present in the original reference). 
If the \textit{Verifier} determines that the report sampled from the \textit{Generator} either does not simulate the target variation correctly or alters the original clinical interpretation, then it rejects the sample, and the \textit{Generator} is prompted again to sample a different report. 
This process is repeated until either the \textit{Verifier} accepts the generated report or a max number of iterations (default: 5) is reached.
In the latter case, we determine that an alternative reference cannot be generated reliably for the given target variation. 
We instantiate the \textit{Generator} and \textit{Verifier} with o3 \citep{o3}, under a setup compliant with the data use agreements for all of the datasets that we consider in our evaluations (Section~\ref{sec:experiments}). 
We provide the remaining details (e.g., prompts, inference setup) in Appendix~\ref{appendix:reref}.

\subsection{Radiologist Validation of \textsc{ReRef}}
\label{sec:annotation}
\vspace{-5pt}

To ensure that \textsc{ReRef} produces alternative reports that are (i) \textit{realistic} (i.e., plausibly written by a practicing radiologist) and (ii) \textit{clinically equivalent to the original report} (i.e., no change in diagnosis and patient management), we conduct a human annotation study to assess each aspect. 
We obtain our annotations from 13 board-certified radiologists---with specialties in cardiothoracic imaging, general radiology, abdomen/pelvis imaging, and neuroradiology---who routinely interpret CXR images. 
For each example, we first obtain annotations from two radiologists. 
If the two annotations disagree, we query a third radiologist to provide their own independent annotation.
If all three annotators disagree, we query a fourth (cardiothoracic imaging) radiologist to provide the final ``consensus'' annotation, after accounting for the rationales provided by the first three annotators. 
We briefly describe each task below and include the detailed annotation guide provided to the radiologists in Appendix~\ref{appendix:annotation}.

\vspace{-7pt}
\paragraph{Task 1: Realism.} We randomly provide either the original radiologist-written report or its alternative version (from \textsc{ReRef}) and ask each annotator to provide one of three scores (1--3): 
\textbf{Realistic (3)}---Reads as a report written by a practicing radiologist in routine clinical care; 
\textbf{Minor Issues (2)}---Mostly realistic, but contains subtle phrasing or formatting choices that are slightly unusual (though clinically correct); 
and \textbf{Major Issues (1)}---Contains major errors, incoherent phrasing, or unnatural elements that a practicing radiologist would not write in a standard radiology report. 
We obtain realism ratings for both the original and alternative reports to test whether the realism ratings for the alternative reports exhibit significant degradation from those of the original reports. 

\vspace{-7pt}
\paragraph{Task 2: Clinical Equivalence.} We provide a randomly sampled (original, alternative) report pair and ask each annotator to provide one of three possible ratings: 
\textbf{Equivalent}---A treating physician reading either report would identify the same findings, and reach the same clinical decisions; 
\textbf{Minor Discrepancy}---Differences between the reports are limited to style, structure, terminology, or level of descriptive detail, and \textit{do not affect what clinical actions would follow}; 
and \textbf{Major Discrepancy}---The reports differ in major ways that could lead a treating physician to different clinical decisions. 

We conduct our study using full radiology reports from four open-access CXR datasets (Section~\ref{sec:experiments}) by subsampling 10 (original, alternative) report pairs per (dataset, axis-aligned variation) combination, resulting in $4 \times 12 \times 10 = 480$ total pairs being annotated. 
We release the subset of 120 annotated MIMIC-CXR report pairs as \textsc{MIMIC-CXR-Ext-ReRef}, excluding others with license restrictions. 

\section{Assessing the Impact of Reporting Practice Variation on RRG Evaluation}
\label{sec:experiments}
To investigate how reporting practice variations impact the conclusions from standard single-reference evaluation of RRG models, we use our \textsc{ReRef} method (Section~\ref{sec:reref}) to perturb the reference report for each imaging study along each axis of our taxonomy (Section~\ref{sec:taxonomy}),
and quantify the resulting change in model performance  
($\text{score}_{\text{alternative}} - \text{score}_{\text{original}}$) as measured by an RRG evaluation metric. 
For each imaging study, we generate \textit{at most} 12 alternative references---each corresponding to an axis-aligned perturbation---while skipping perturbations that are infeasible for the original reference provided (e.g., without hallucinating). 

We focus on the task of generating the \textit{Findings} section of each radiology report, which is most common for RRG evaluation\footnote{For instance, the ReXrank leaderboard \citep{rexrank} for Findings generation includes 29 models, while that for Findings \& Impression generation only includes 9 models (as of Sep. 9, 2026).}. 
For a given imaging study, each RRG model generates the Findings section based on the images and context (Indication, Comparison, and Technique sections) from the \textit{current exam only}. 
We leave longitudinal settings, in which images and/or reports from prior exams are also provided as input \citep{med-cxrgen,maira-2,lunguage}, to future work. 

Notably, for perturbations along the \textit{Section Assignment} axis, which reorganize clinical findings between the Findings and Impression sections, the Findings-only scoping can amplify the measured sensitivity of metrics. 
We nonetheless retain this setup to reveal the ways in which current evaluation practices are sensitive to reporting practice variations in the reference. 
Below, we provide additional details on the datasets, models, and evaluation metrics used.

\vspace{-7pt}
\paragraph{Datasets.} We consider four widely used open-access CXR report datasets for evaluation: MIMIC-CXR~\citep{mimic-cxr}, CheXpert Plus~\citep{chexpertplus}, IU X-ray~\citep{iu-xray}, and ReXGradient-160K~\citep{rexgradient}. 
For MIMIC-CXR, we use the official test set on PhysioNet \citep{physionet} (3269 studies).
For CheXpert Plus, we use the official validation set (200 studies), as the test set is not publicly available.
For IU X-ray, we use the 590 imaging studies from the test split provided by \citet{r2gen}, following the ReXrank leaderboard \citep{rexrank}.
For ReXGradient-160K, we use the official \textit{public} test set (10000 studies). 
From these test sets, we exclude any studies without a Findings section to be used as the reference (Appendix~\ref{appendix:datasets}). 
Table~\ref{tab:dataset-summary} shows the final number of examples from each dataset used for analysis. 

\vspace{-7pt}
\paragraph{Models.} We evaluate 9 open-source RRG models: MAIRA-2 \citep{maira-2}, MedGemma-4B \citep{medgemma}, CheXOne \citep{chexone}, MedVersa \citep{medversa}, CheXagent-2-3B \citep{chexagent}, Med-CXRGen \citep{med-cxrgen}, Libra \citep{libra}, LLaVA-Rad \citep{llava-rad}, and CheX-MIMIC \citep{chexpertplus}. 
By CheX-MIMIC, we refer to the baseline model from \citet{chexpertplus} which was trained on both CheXpert Plus and MIMIC-CXR. 
We select these models to cover a wide range of model architectures and sizes, focusing on those that perform strongly in prior works \citep{rexrank}. 
For models trained to handle multi-view CXR image inputs (e.g., frontal and lateral), we provide all views available for each imaging study, as feasible within the model's constraints. 
Otherwise, we only provide the frontal view as input.
We provide the remaining details (e.g., prompts, inference setup) in Appendix~\ref{appendix:models}.

\vspace{-7pt}
\paragraph{Evaluation Metrics.} We consider the following RRG evaluation metrics: (i) LLM judge metrics: CRIMSON \citep{crimson}, GREEN \citep{green}; (ii) clinical entity-based metrics: RadGraph-F1 \citep{radgraph,radgraph-f1}, RaTEScore \citep{ratescore}; (iii) composite metrics: RadCliQ-v1 \citep{radcliq}; (iv) disease label-based metrics: CheXbert-F1 \citep{chexbert}; and (v) NLG metrics: BLEU-2 \citep{bleu}, ROUGE-L \citep{rouge}, BERTScore \citep{zhang2020bertscoreevaluatingtextgeneration}.
For CRIMSON and GREEN, we use the open-source implementations based on MedGemma-4B \citep{medgemma} and Llama-2-7B \citep{llama-2}, respectively, as provided by the authors. 
For CheXbert-F1, we compute the micro-average F1 scores based on 5 out of the 14 supported disease labels---atelectasis, cardiomegaly, consolidation, edema, and pleural effusion---following prior works \citep{radgraph-f1}.
All metrics range between 0 and 1, \textit{except} CRIMSON (range: [-1,1]) and RadCliQ-v1 (range: (0,$\infty$)). 
As RadCliQ-v1 is the only metric where lower is better, we flip its sign unless specified otherwise, to match the behavior of the other metrics.
While taking the reciprocal (i.e., 1/RadCliQ-v1) is more standard in the literature \citep{rexrank,unirg-cxr}, we avoid doing so, as it becomes unstable when the raw values approach 0. 

\section{Results}
\label{sec:results}
\begin{figure*}[t!]
    \centering
    \begin{tabular}{@{}c@{}c@{}}
        \begin{subfigure}{0.03\linewidth}
            \makebox[\linewidth]{\raisebox{90pt}{{(a)}}} 
        \end{subfigure} &
        \begin{subfigure}{0.9\linewidth} 
            \includegraphics[width=\linewidth]{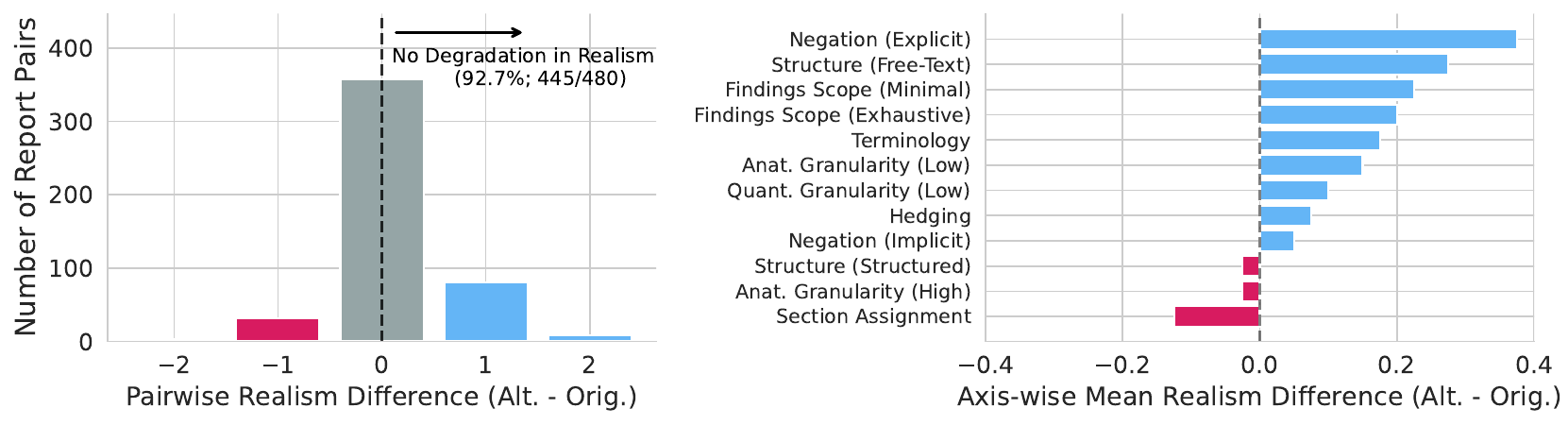}
        \end{subfigure} \\
        \begin{subfigure}{0.03\linewidth}
            \makebox[\linewidth]{\raisebox{90pt}{{(b)}}} 
        \end{subfigure} &
        \begin{subfigure}{0.9\linewidth} 
            \includegraphics[width=\linewidth]{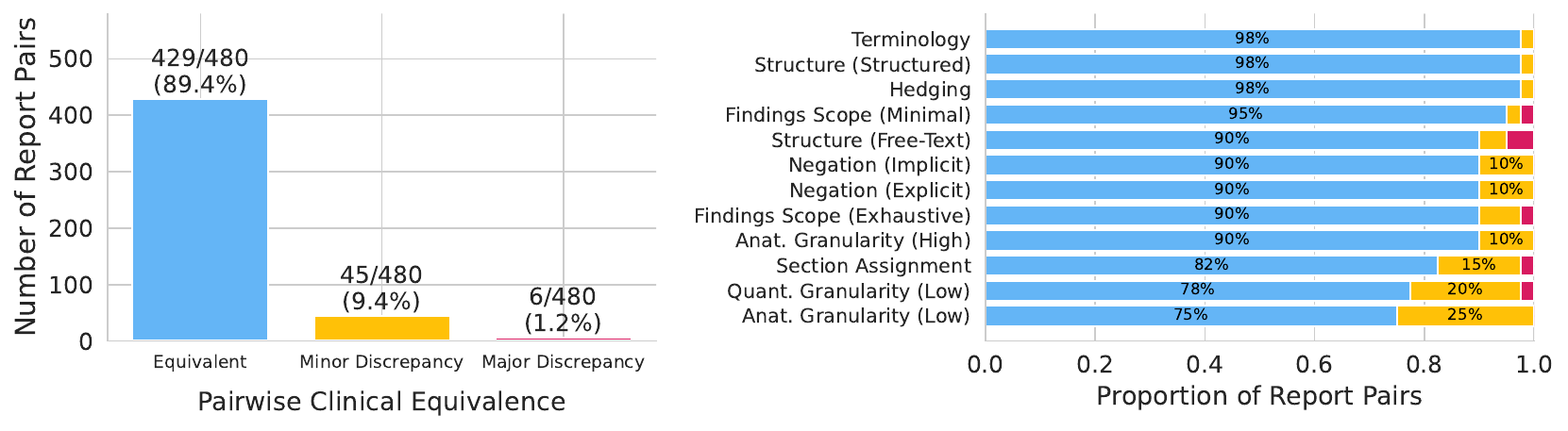}
        \end{subfigure}
    \end{tabular}
    \vspace{-8pt}
    \caption{Radiologists find that \textsc{ReRef} generates alternative references that are realistic and preserve the original clinical interpretation (Section~\ref{sec:results-reref}). 
    (a) Task 1 (Realism): In 92.7\% of all report pairs, the generated alternative reference is deemed as realistic as the original reference (left). 
    (b) Task 2 (Clinical Equivalence): In 98.8\% of all report pairs, the generated alternative reference is labeled as preserving the clinical interpretation without change in clinical impact (\textit{Equivalent} or \textit{Minor Discrepancy}) (left).
    In both tasks, trends vary across different axis-aligned perturbations (right).
    }
    \label{fig:reref-results}
    \vspace{-5pt}
\end{figure*}

Here, we summarize the main findings from our experiments. 
Unless specified otherwise, we focus our discussion on the following metrics that prior works suggest to correlate well with radiologist judgment and are widely used: CRIMSON, GREEN, RadCliQ-v1, RaTEScore, and RadGraph-F1.
\vspace{-5pt}

\subsection{\textsc{ReRef} Generates Realistic Alternative References That Preserve Clinical Interpretation}
\label{sec:results-reref}
\vspace{-5pt}

From the expert annotation tasks outlined in Section~\ref{sec:annotation}, 
we find that radiologists generally find \textsc{ReRef} to be effective in generating realistic alternative references without modifying the original clinical interpretation (Figure~\ref{fig:reref-results}). 
For Task 1 (Realism), the alternative report is labeled to be as realistic as the original (i.e., $\text{realism}_{\text{alternative}} - \text{realism}_{\text{original}} \geq 0$) in 92.7\% (445 out of 480) of the annotated (original, alternative) report pairs (Figure~\ref{fig:reref-results}(a, left)). 
Of the $480 \times 2 = 960$ individual reports reviewed, radiologists unanimously agree on the realism rating in 567 cases (59.1\%), achieve majority agreement in 345 cases (35.9\%), and reach a three-way disagreement (resolved by a fourth annotator; Section~\ref{sec:annotation}) in 48 cases (5.0\%). 
The realism trends vary across different perturbations---e.g., \textit{Section Assignment} leads to a slight degradation, while \textit{Findings Scope (Minimal)} and \textit{Anatomical Granularity (Low)} do not---but 9 out of 12 achieve no degradation on average 
(Figure~\ref{fig:reref-results}(a, right)).

For Task 2 (Clinical Equivalence), only 1.2\% of all report pairs (6 out of 480) are given the \textit{Major Discrepancy} rating (\textit{Equivalent}: 89.4\%; \textit{Minor Discrepancy}: 9.4\%), indicating that \textsc{ReRef} generally preserves the original clinical content without altering the implied diagnosis and patient management (Figure~\ref{fig:reref-results}(b, left)). 
Of the 480 report pairs reviewed, radiologists unanimously agree on the equivalence rating in 329 cases (68.5\%), achieve majority agreement in 140 cases (29.2\%), and reach a three-way disagreement (resolved by a fourth annotator) in 11 cases (2.3\%).
The specific trends vary across different perturbations, with some (notably, those along the \textit{Granularity} dimension) exhibiting higher \textit{Minor Discrepancy} rates 
(Figure~\ref{fig:reref-results}(b, right)). 
Meanwhile, we find that radiologists often use \textit{Minor Discrepancy} to mark changes in the level of detail that they deem clinically inconsequential (e.g., ``Report A is slightly more specific in location... a treating physician will notice without changing clinical management.'')---exactly the kind of variation these perturbations are designed to introduce. 

\begin{figure*}[t!]
    \centering
    \begin{tabular}{@{}c@{}}
        \begin{subfigure}{0.9\linewidth} 
            \includegraphics[width=\linewidth]{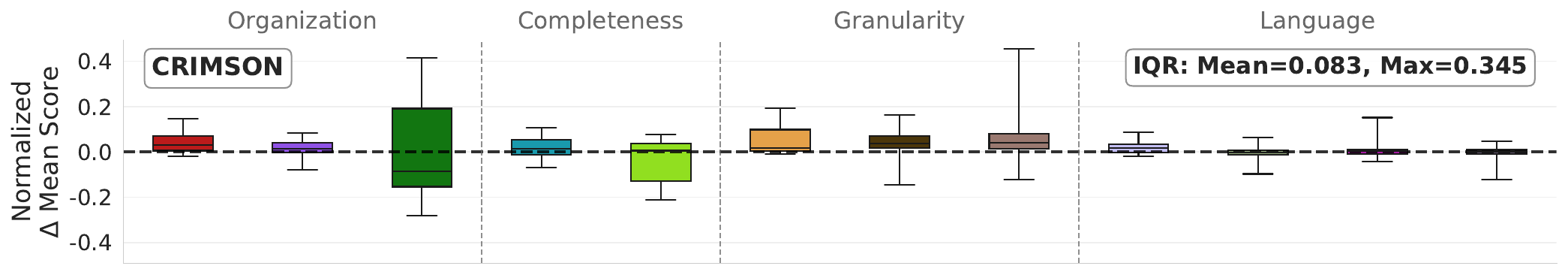}
        \end{subfigure} \\
        \begin{subfigure}{0.9\linewidth}
            \includegraphics[width=\linewidth]{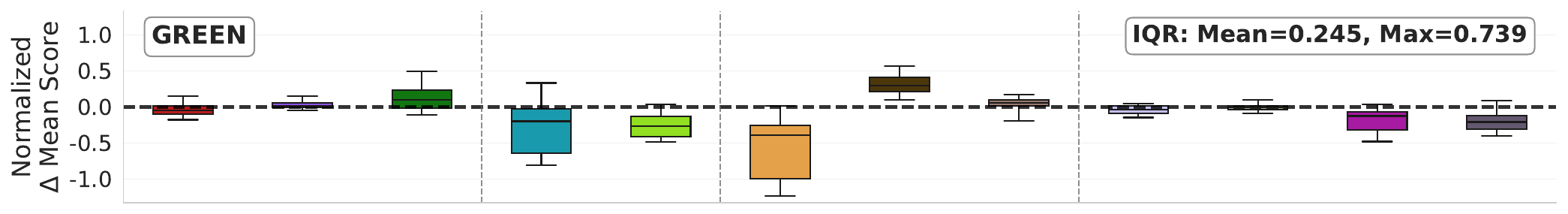}
        \end{subfigure} \\
        \begin{subfigure}{0.9\linewidth}
            \includegraphics[width=\linewidth]{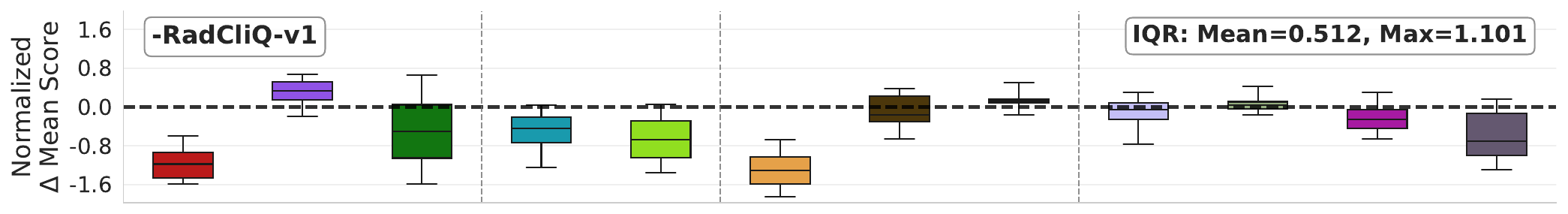}
        \end{subfigure} \\
        \begin{subfigure}{0.9\linewidth}
            \includegraphics[width=\linewidth]{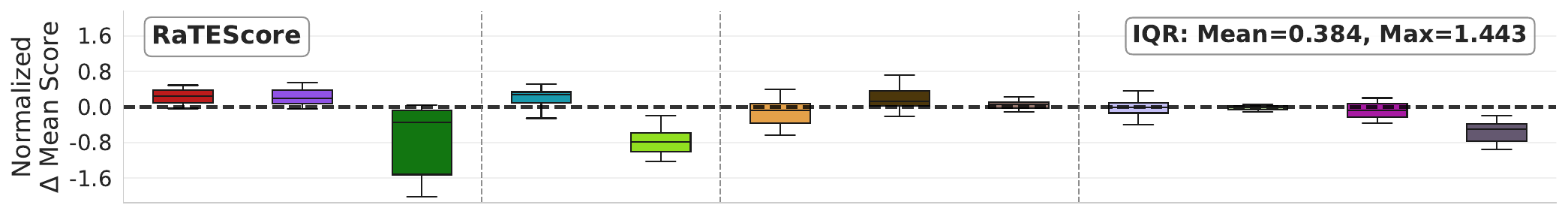}
        \end{subfigure} \\
        \begin{subfigure}{0.9\linewidth}
            \includegraphics[width=\linewidth]{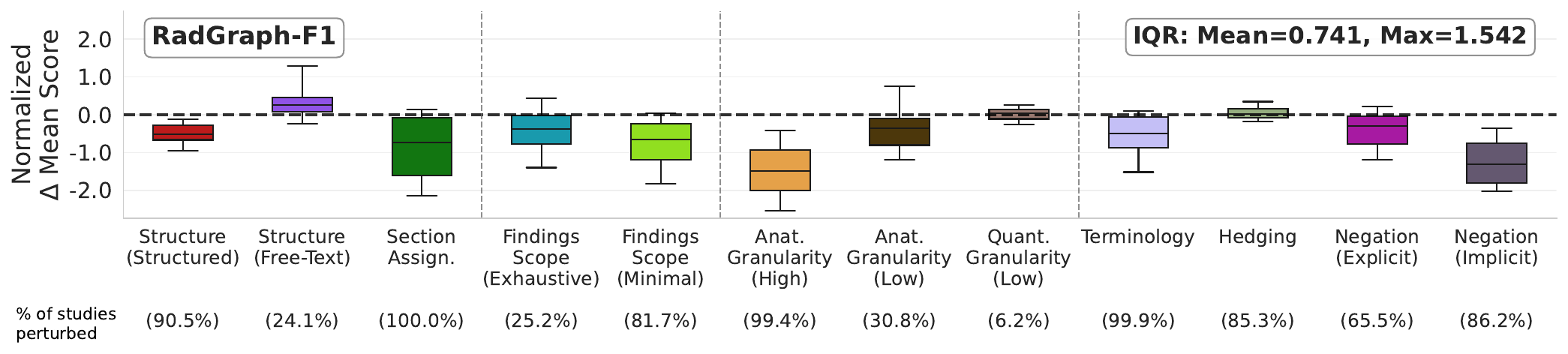}
        \end{subfigure}
    \end{tabular}
    \vspace{-7pt}
    \caption{Most of the evaluation metrics we consider are sensitive to reporting practice variations in the reference (Section~\ref{sec:results-metric-sensitivity}).
    Each box plot shows the distribution, across $9 \times 4 = 36$ (model, dataset) pairs, of changes in the average model scores
    induced by each axis-aligned perturbation, 
    normalized by the metric's standard deviation across (model, dataset) pairs under the original references. 
    The lower and upper whiskers correspond to the 5th and 95th percentiles of the normalized differences, respectively. 
    For each metric, the \textit{mean} and \textit{max} interquartile range (IQR) values, aggregated over the different axes, indicate the \textit{overall} and \textit{worst-case} sensitivity of the metric to reporting practice variations. 
    Score changes are computed over the subset of studies for which each perturbation is feasible (Section~\ref{sec:experiments}), whose proportion we report in parentheses below each $x$-axis label.
    }
    \label{fig:global-metric-sensitivity}
    \vspace{-5pt}
\end{figure*}

\subsection{
Measured Clinical Accuracy Can Depend on the Reporting Practices of the Reference
}
\label{sec:results-metric-sensitivity}
\vspace{-5pt}

\begin{figure*}[t!]
    \centering
    \includegraphics[width=0.95\linewidth]{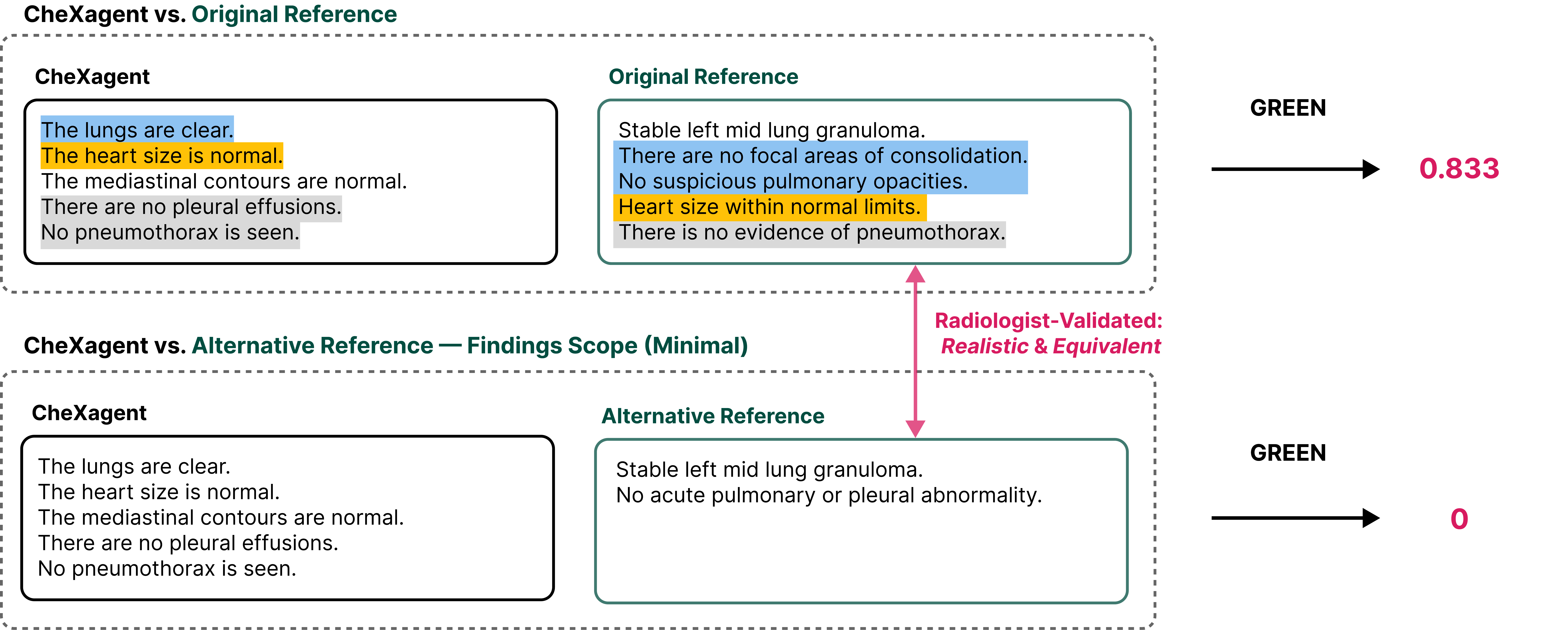} 
    \vspace{-2pt}
    \caption{Assessment of clinical accuracy can be impacted by reporting practices of the reference (Section~\ref{sec:results-metric-sensitivity}). 
    We show an example based on the IU X-ray dataset, where a report generated by CheXagent is individually compared using GREEN against a radiologist-validated (original, alternative) report pair generated by a \textit{Findings Scope (Minimal)} perturbation using \textsc{ReRef} (Section~\ref{sec:reref}). 
    Findings highlighted with the same color indicate the ``matched'' findings, as determined by GREEN. 
    Under the alternative reference, a mismatch in the granularity of the normal pleural findings causes the CheXagent score to drop to 0, despite partial agreement in clinical interpretation.
    }
    \label{fig:metric-sensitivity-example}
    \vspace{-5pt}
\end{figure*}

Under targeted perturbations to the reference report along each axis of our taxonomy (Section~\ref{sec:experiments}), 
we find that many metrics shift substantially in response to reporting practice variations (Figure~\ref{fig:global-metric-sensitivity})---shifts that, as we show below, cannot be attributed to clinical content alone
(Figure~\ref{fig:metric-sensitivity-example}). 
In Figure~\ref{fig:global-metric-sensitivity}, each box plot shows the distribution, across $9 \times 4 = 36$ (model, dataset) pairs, of changes in the average model scores
$\frac{1}{n}\sum_{i=1}^n (\text{score}^{(i)}_{\text{alternative}} - \text{score}^{(i)}_{\text{original}})$ induced by each axis-aligned perturbation, 
normalized by the metric's standard deviation across (model, dataset) pairs under the original references so that the shifts are interpretable on the scale of the metric's natural variation. 
The lower and upper whiskers correspond to the 5th and 95th percentiles, respectively. 
We also report the mean and max interquartile range (IQR) values (aggregated across all axis-aligned perturbations) as summary statistics of the \textit{overall} and \textit{worst-case} sensitivity of each metric to reporting practice variations. 

CRIMSON emerges as the most stable metric overall (mean IQR: 0.083), 
but still exhibits noticeable variations along certain axes (max IQR: 0.345), such as \textit{Findings Scope (Minimal)} and \textit{Section Assignment}. 
Compared to CRIMSON, GREEN exhibits higher overall and worst-case sensitivity (mean IQR: 0.245, max IQR: 0.739), partially due to its additional sensitivity to any changes to the \textit{normal} findings in the reports, which CRIMSON ignores by design \citep{crimson}. 
The remaining non-LLM-based metrics---RadCliQ-v1 (IQR: mean = 0.512, max = 1.101), RaTEScore (IQR: mean = 0.384, max = 1.443), and RadGraph-F1 (IQR: mean = 0.741, max = 1.542)---exhibit substantially higher sensitivity, with some perturbations leading to shifts in the average model scores that are 1.5--2 times as large as each metric's natural variation (e.g., RadCliQ-v1 in response to \textit{Structure (Structured)}, RadGraph-F1 in response to \textit{Anatomical Granularity (High)}). 

While some sensitivity to the details of the reference is expected, 
we find that reporting practice, rather than clinical content, can drive such score changes. 
To illustrate, we show an example from IU X-ray, where a report generated by CheXagent is individually compared against a radiologist-validated (original, alternative) report pair using GREEN, under the \textit{Findings Scope (Minimal)} perturbation for which \textsc{ReRef} tends to be more reliable (Figure~\ref{fig:reref-results}, right).
In both cases, CheXagent misses the reference finding of ``Stable left mid lung granuloma'' but correctly identifies the absence of other pleural abnormalities. 
Against the original reference, GREEN credits CheXagent for this normal interpretation, resulting in a score of 0.833.
However, against the alternative reference, \textbf{the same prediction receives no credit and leads to a score of 0}, as the \textit{Findings Scope (Minimal)} perturbation condenses the reference's pleural normal findings into a higher-level summary that GREEN no longer treats as a match.
The two references are clinically equivalent, yet the score collapses from 0.833 to 0, and the entire score difference is attributable to how the reference chooses to express the normal findings. 
Such behavior illustrates the sensitivity of metrics like GREEN to the reporting practices of the reference, beyond the true accuracy of the underlying clinical interpretations. 

\subsection{Reporting Practice Variations Can Alter Conclusions from Model Comparison}
\label{sec:results-nonuniform-impact}
\vspace{-5pt}

\begin{figure*}[t!]
    \centering
    \includegraphics[width=0.95\linewidth]{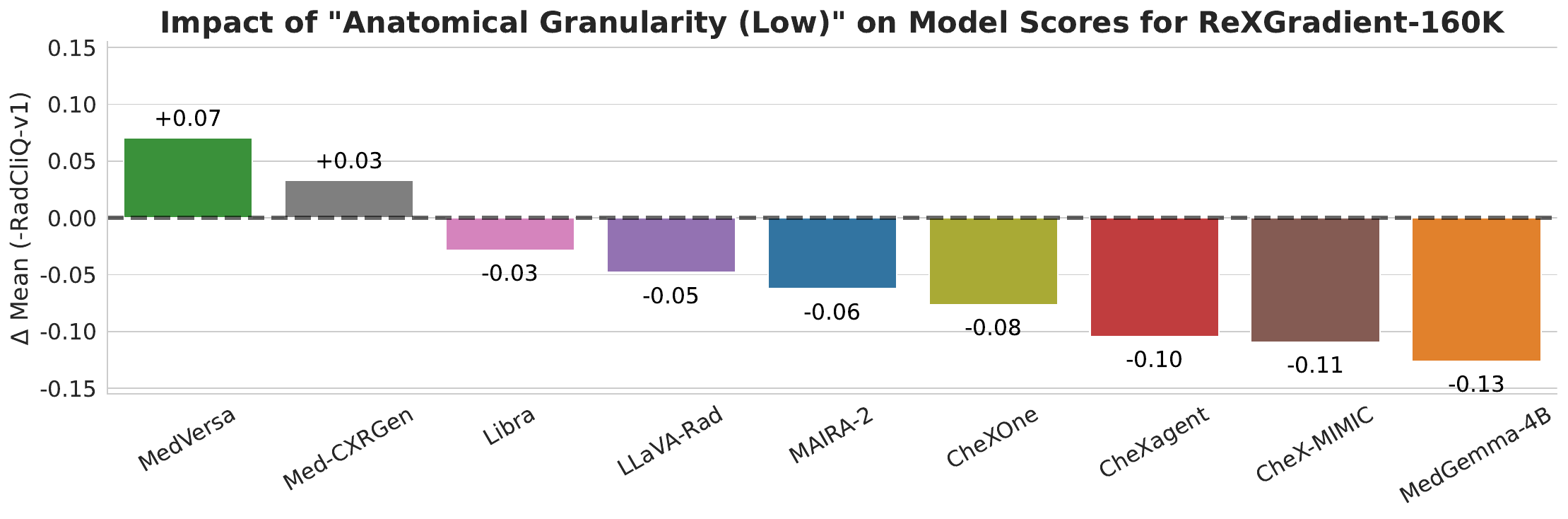} 
    \vspace{-7pt}
    \caption{Axis-aligned perturbations to the reference reporting practice can impact model scores unevenly (Section~\ref{sec:results-nonuniform-impact}). On ReXGradient-160K, the \textit{Anatomical Granularity (Low)} perturbation, which merges separate anatomical mentions (e.g., ``heart'', ``mediastinal'') into compound anatomical findings (e.g., ``cardiomediastinal''), tends to advantage models like MedVersa (+0.07), which exhibit a similar tendency, over others like MedGemma-4B (-0.13) that exhibit the opposite tendency. 
    The average model scores are computed based on the subset of studies in ReXGradient-160K for which the \textit{Anatomical Granularity (Low)} perturbation is feasible ($\sim$29.3\%) (Section~\ref{sec:experiments}).
    }
    \label{fig:nonuniform-impact}
    \vspace{-4pt}
\end{figure*}

We also find that the sensitivity of metrics to reporting practice can unevenly impact the scores of different models (Figure~\ref{fig:nonuniform-impact}). 
For example, under RadCliQ-v1, 
\textit{Anatomical Granularity (Low)} systematically benefits certain models over others on ReXGradient-160K, resulting in positive score changes for MedVersa (+0.07) and Med-CXRGen (+0.03), and negative score changes for others like MedGemma-4B (-0.13) and CheX-MIMIC (-0.11). 
These trends emerge as \textit{Anatomical Granularity (Low)} merges separate anatomical mentions into compound anatomical findings (e.g., ``The \textbf{heart size} and \textbf{mediastinal} contours are within normal limits.'' $\rightarrow$ ``Normal \textbf{cardiomediastinal} silhouette.''). Such a change leads to a higher number of matches in clinical entities for models like MedVersa, which is relatively terse in its reports and exhibits a similar tendency (e.g., uses ``cardiomediastinal'' in 59.6\% of all studies in ReXGradient-160K), in contrast to others like MedGemma-4B that tend to enumerate the lower-level anatomical structures (e.g., uses ``cardiomediastinal'' in only 0.7\% of all studies in ReXGradient-160K). 
As RadCliQ-v1 is computed based on metrics like RadGraph-F1 that require exact clinical entity matches, these changes disparately impact different models.

\begin{figure*}[t!]
    \centering
    \begin{tabular}{@{}c@{}c@{}c@{}c@{}}
         \begin{subfigure}{0.03\linewidth}
             \makebox[\linewidth]{\raisebox{70pt}{{(a)}}} 
         \end{subfigure} &
         \begin{subfigure}{0.95\linewidth} 
            \includegraphics[width=\linewidth]{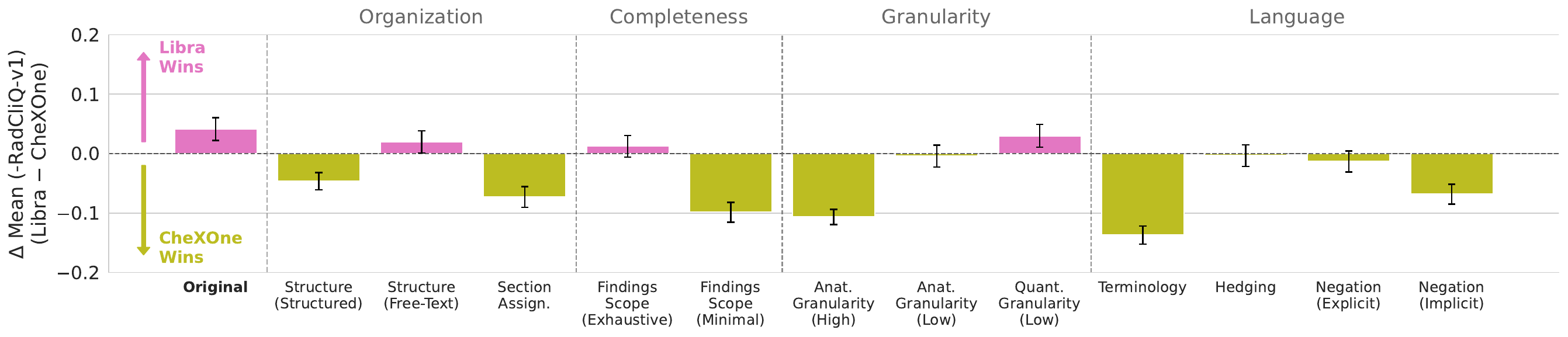}
         \end{subfigure} \\
         \begin{subfigure}{0.03\linewidth} 
             \makebox[\linewidth]{\raisebox{70pt}{{(b)}}} 
         \end{subfigure} &
         \begin{subfigure}{0.95\linewidth} 
            \includegraphics[width=\linewidth]{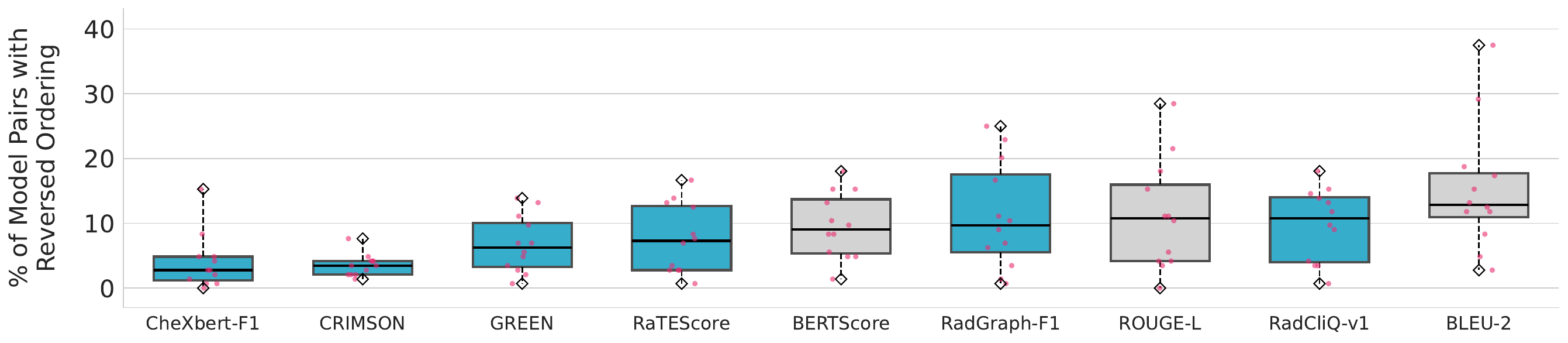}
         \end{subfigure}
    \end{tabular}
    \caption{Model rankings are sensitive to the reporting practices of the reference (Section~\ref{sec:results-nonuniform-impact}). 
    (a) On MIMIC-CXR, the relative ordering of Libra~\citep{libra} vs.~CheXOne~\citep{chexone} based on average score varies across different axis-aligned perturbations. 
    Each bar displays the difference in average RadCliQ-v1 ($\text{Libra} - \text{CheXOne}$), where the sign of RadCliQ-v1 has been flipped so that higher is better (Section~\ref{sec:experiments}), and the error bars denote the 95\% bootstrap confidence intervals (CIs) in this difference. 
    When a perturbation is infeasible for an imaging study, we retain its original-reference score, so the average scores are always computed using all studies. 
    (b) Aggregating across all datasets and model pairs, we find that for all metrics, the model rankings can change after perturbation, albeit to varying extents. 
    Each box plot shows, across the 12 axis-aligned perturbations, the percentage of all $4 \times \binom{9}{2} = 144$ (dataset, model pair) comparisons whose ordering by mean score reverses relative to the original reference, with the diamond-shaped markers denoting the max and min values across all perturbations. 
    The cyan-colored box plots indicate the clinical metrics.
    }
    \label{fig:ranking-sensitivity}
    \vspace{-5pt}
\end{figure*}

\textbf{Ultimately, we find that such changes can 
alter the broader conclusions of RRG model comparison} (Figure~\ref{fig:ranking-sensitivity}). 
For example, in Figure~\ref{fig:ranking-sensitivity}(a), we show the gap in average RadCliQ-v1 between Libra and CheXOne on MIMIC-CXR, under the original (left) vs.~alternative references (middle--right), where we consider a pairwise comparison to be \textit{significant} if the 95\% bootstrap confidence interval (CI) for this gap\footnote{We compute the 95\% bootstrap CIs by resampling the test set with replacement 10,000 times.} excludes zero, and a statistical \textit{tie} otherwise. 
When a perturbation is infeasible for a study, we retain its original-reference score, and always compute the average scores using all studies. 
Under the original references, Libra significantly outperforms CheXOne. 
However, after the perturbations, 
Libra significantly \textit{underperforms} CheXOne in 6 settings (e.g., \textit{Findings Scope (Minimal)}), reaches a tie in 4 settings (e.g., \textit{Anatomical Granularity (Low)}), and significantly outperforms it in only 2 settings (e.g., \textit{Structure (Free-Text)}). 

Aggregating the results across all datasets and model pairs, we find that for all metrics, the model rankings can deviate from those obtained under the original references, 
in response to the perturbations (Figure~\ref{fig:ranking-sensitivity}(b)). 
For 7 out of 9 metrics, such changes remain noticeable even when we limit to pairwise ranking reversals whose 95\% bootstrap CI-based comparisons change (Figure~\ref{fig:ranking-sensitivity-significant}), with CRIMSON and CheXbert-F1 being the exceptions (Appendix~\ref{appendix:ranking-reversals}). 
Notably, rankings under RadCliQ-v1, the default metric for leaderboards like ReXrank \citep{rexrank}, 
show the highest \textit{median} sensitivity to changes in reporting practice among all clinical metrics (Figure~\ref{fig:ranking-sensitivity}(b); in cyan). 

Meanwhile, while we focus on single-axis perturbations in our main experiments, real-world radiology reports likely reflect \textit{multiple} reporting practices simultaneously and in a \textit{correlated} manner (e.g., \textit{Findings Scope (Minimal)} + \textit{Anatomical Granularity (Low)}). We find that compounding multiple perturbations can lead to even more disruptions to the model rankings (Figures~\ref{fig:ranking-sensitivity-compound-minimalist}--\ref{fig:ranking-sensitivity-compound-maximalist}; Appendix~\ref{appendix:compound-perturbations}). 

\section{Conclusion}
\label{sec:conclusion}
In this work, we demonstrated that many automated metrics for RRG model evaluation---including those reported to correlate well with radiologist judgment---are sensitive to variations in the reporting practices of the reference, 
to the extent that substantive conclusions about the performance of RRG models can meaningfully change. 
In surfacing these findings, we (i) introduced a taxonomy that characterizes the various ways in which radiologists vary in how they communicate the same clinical interpretation; (ii) developed \textsc{ReRef}, an expert-validated method for rewriting reference reports to align with specific reporting practices;
and (iii) conducted extensive evaluations with nine RRG models and four CXR datasets to quantify the impact of each variation on model performance. 

Our findings suggest that many current metrics may fail to decouple clinical interpretation from conformity to reporting practices of the reference. 
Thus, choosing the ``right'' references that best-reflect the desired reporting practices for the setting of interest can be an important design decision for model evaluation. 
For instance, when assessing potential deployment at an institution, the reference reports should reflect local reporting conventions, since adherence to those conventions may itself be an integral part of report quality. 
When assessing a model's ability to document fine-grained clinical findings, the references should exhibit sufficient \textit{Completeness} and \textit{Granularity}, as a sparse, abstract reference cannot be used to verify the accuracy of more detailed findings generated by a model.

For broader model comparisons, we recommend using our \textsc{ReRef} method to simulate and evaluate against multiple clinically equivalent references spanning practically relevant reporting practices and reporting the resulting variability in model scores and rankings. 
The objective is not for every model to receive the same score under every reference, but to ensure that the positive conclusions about a model's effectiveness do not depend on an arbitrary choice of reporting practice in the reference. 

\vspace{-5pt}
\paragraph{Limitations.} While we conduct extensive radiologist validation to verify \textsc{ReRef}'s capability in generating realistic and clinically equivalent alternative references, it is possible that the generated outputs exhibit hallucinations or subtle deviations in clinical interpretation. 
As \textsc{ReRef} is implemented based on a proprietary, closed-source LLM (o3), reproducing the alternative references used in our perturbation experiments may be difficult. 
Nonetheless, our shared repository contains the complete generation and verification pipeline used in our experiments, and all of the prompting details are included in Appendix~\ref{appendix:reref}. 
As LLM judge metrics (CRIMSON, GREEN) can be implemented with any LLM of choice, the specific trends observed in our study may change when a different model is used. 

\vspace{-5pt}
\section*{Acknowledgments and Disclosure of Funding}
We gratefully acknowledge DARPA (FA8750-23-2-1015), ONR (N00014-23-1-2368), NSF (IIS-2211955), UPMC, Highmark Health, Abridge, Ford Research, Mozilla, the PwC Center, Amazon AI, JP Morgan Chase, the Block Center, the Center for Machine Learning and Health, and the CMU Software Engineering Institute (SEI) via Department of Defense contract FA8702-15-D-0002, for their generous support of our research.


\bibliography{refs}

\newpage


\appendix

\renewcommand\thetable{A\arabic{table}}
\renewcommand\thefigure{A\arabic{figure}}
\setcounter{table}{0}
\setcounter{figure}{0}
\section{Additional Details on \textsc{ReRef}}
\label{appendix:reref}
Here, we provide additional details for \textsc{ReRef} (Section~\ref{sec:reref}).
We first describe the precise definitions, allowed variations, illustrative examples, and axis-specific constraints used to operationalize each axis of our taxonomy (Appendix~\ref{appendix:reref-axes}).
We then describe the prompts used to instantiate the \textit{Generator} and \textit{Verifier} (Appendix~\ref{appendix:reref-prompts}), as well as the inference setup used to deploy them (Appendix~\ref{appendix:reref-inference}).

\subsection{Axis-Aligned Perturbations: Definitions, Examples, and Constraints}
\label{appendix:reref-axes}

For each axis of our taxonomy (Section~\ref{sec:taxonomy}), we specify its  (i)~\textit{description}, (ii)~a set of allowed \textit{variations} (each with a short definition), (iii)~illustrative \textit{examples} contrasting the variations carefully designed with a board-certified radiologist, and (iv)~zero or more axis-specific \textit{constraints} that supplement the default constraints listed in Appendix~\ref{appendix:reref-prompts}.
To operationalize perturbations along each axis, we further distinguish between axes whose variations can be characterized in a standalone manner (e.g., for the \textit{Structure} axis, a reference is either \textit{Structured} or \textit{Free-Text})---which we refer to as \textit{absolute} axes---and axes whose variations are best characterized relative to the original reference (e.g., \textit{Terminology} is defined as a near-synonymous rewording of the original)---which we refer to as \textit{relative} axes.
For the absolute axes, the \textit{Generator} produces alternative references for every variation other than the one that the original reference is already in (e.g., both \textit{Structure (Structured)} and \textit{Structure (Free-Text)} are sampled, with infeasibility handled at the \textit{Verifier} stage). 
For the relative axes, we simply prompt the LLM to sample an alternative version that differs from the original reference along the given axis. 
The descriptions, allowed variations, examples, and constraints below are provided to the \textit{Generator} and \textit{Verifier} as part of the prompts shown in Appendix~\ref{appendix:reref-prompts}.

\subsubsection{Organization: Structure (Absolute)}
\textbf{Description:} Radiologists differ in their use of structured templates versus continuous free-text prose. Some use highly templated formats (e.g., explicit headings for each anatomical region, bullet points, numbered lists, subheadings for groups of findings), while others write in continuous free-text sentences and paragraphs.

\textbf{Variations:}
\begin{itemize}\itemsep0pt
    \item \textit{Structured}: Written in highly structured templated format (e.g., explicit headings for each anatomical region, bullet points, numbered lists, subheadings for groups of findings).
    \item \textit{Free-Text}: Written in continuous free-text prose.
\end{itemize}

\textbf{Example 1:}
\begin{itemize}
    \item \textit{Structured}: ``CARDIOMEDIASTINAL: Normal silhouette. LUNGS: Well expanded and clear. PLEURA: Sharp costophrenic sulci and diaphragmatic contour. BONES: No acute abnormalities.''
    \item \textit{Free-Text}: ``The cardiomediastinal silhouette is normal. The lungs are well expanded and clear. The costophrenic sulci and diaphragmatic contour are sharp. No acute osseous abnormalities.''
\end{itemize}

\textbf{Example 2:}
\begin{itemize}
    \item \textit{Structured}: ``1.~Single view of chest demonstrates no significant interval change in position of bilateral chest tubes. Median sternotomy wires are unchanged. 2.~Previously described left pneumothorax is not well seen on today's exam. No large pneumothorax is seen. 3.~Interval increase in left pleural effusion. Small right pleural effusion present. Summary: Possible significant abnormality and change. May need action.''
    \item \textit{Free-Text}: ``Single view of chest demonstrates no significant interval change in position of bilateral chest tubes. Median sternotomy wires are unchanged. Previously described left pneumothorax is not well seen on today's exam. No large pneumothorax is seen. Interval increase in left pleural effusion. Small right pleural effusion present. Possible significant abnormality and change. May need action.''
\end{itemize}

\textbf{Example 3:}
\begin{itemize}
    \item \textit{Structured}: ``1.~Cardiomegaly. 2.~Resolution of right basal infiltrate.''
    \item \textit{Free-Text}: ``Cardiomegaly. Resolution of right basal infiltrate.''
\end{itemize}

\textbf{Axis-Specific Constraints:} The Findings and Impression sections need not be structured in the same way; e.g., the Findings section can be more structured while the Impression section is more free-text, and vice versa.

\subsubsection{Organization: Section Assignment (Relative)}
\textbf{Description:} Radiologists differ in where they place specific types of content within the report (Findings vs.~Impression). For example, some discuss prior comparisons only in the Impression section to keep the Findings section fully grounded on the current study, while others discuss them in the Findings section to provide immediate clinical context. Meanwhile, there are also cases where comparison findings are discussed in both sections. 
As another example, some discuss differential diagnoses in both the Findings and the Impression, while others only discuss them in the Impression.

\textbf{Variations:}
\begin{itemize}\itemsep0pt
    \item \textit{Original}: Retains the original assignment of content to the Findings and Impression sections.
    \item \textit{Alternative}: Reassigns content to different sections relative to the original reference.
\end{itemize}

\textbf{Example 1:}
\begin{itemize}
    \item \textit{Original}: ``FINDINGS: Chronic left-sided rib fractures are again noted. The cardiomediastinal and hilar contours are unchanged from \_\_\_. Pleural thickening and blunting at the right costophrenic angle is again demonstrated, and is stable from the prior exam in \_\_\_ and likely represents pleural scarring and a small pleural effusion. No focal consolidation or pneumothorax is identified.~~IMPRESSION: Multiple chronic appearing left-sided rib fractures. No pneumothorax. Blunting of the costophrenic angle on the right likely represents pleural scarring and a small effusion, not significantly changed from \_\_\_.''
    \item \textit{Alternative}: ``FINDINGS: Left-sided rib fractures are present. The cardiomediastinal and hilar contours are normal. Pleural thickening and blunting at the right costophrenic angle, likely representing pleural scarring and a small pleural effusion. No focal consolidation or pneumothorax is identified.~~IMPRESSION: Multiple chronic left-sided rib fractures, again noted and unchanged from \_\_\_. Cardiomediastinal and hilar contours remain stable. No pneumothorax. Blunting of the costophrenic angle on the right likely represents pleural scarring and a small effusion, again demonstrated and stable from prior exam in \_\_\_.''
\end{itemize}

\textbf{Example 2:}
\begin{itemize}
    \item \textit{Original}: ``FINDINGS: A hazy opacity is present in the right lung, which may represent aspiration, pleural effusion or hemorrhage. Moderate cardiomegaly is stable. Slight prominence of the pulmonary vasculature with cephalization and enlarged pulmonary arteries are consistent with mild pulmonary edema. There are no displaced rib fractures.~~IMPRESSION: 1.~Hazy opacity in the right lung, which may represent aspiration versus pleural effusion or hemorrhage. 2.~Mild pulmonary edema. 3.~No displaced rib fractures.''
    \item \textit{Alternative}: ``FINDINGS: A hazy opacity is present in the right lung. Moderate cardiomegaly is stable. Slight prominence of the pulmonary vasculature with cephalization and enlarged pulmonary arteries. There are no displaced rib fractures.~~IMPRESSION: 1.~Hazy opacity in the right lung, which may represent aspiration versus pleural effusion or hemorrhage. 2.~Mild pulmonary edema. 3.~No displaced rib fractures.''
\end{itemize}

\textbf{Axis-Specific Constraints:} The union of clinical content across the Findings and Impression sections must match that of the original reference. The \textit{Generator} must not (i)~introduce details from any other sections (e.g., Technique, Indication, Comparison) or (ii)~omit any clinical details from the original Findings and Impression sections.

\subsubsection{Completeness: Findings Scope (Absolute)}
\textbf{Description:} Radiologists differ in how exhaustively they describe normal and abnormal findings. Some describe all normal and abnormal findings exhaustively, while others selectively omit normal findings and focus on a minimal list of abnormal findings most likely to be clinically relevant to the patient's condition and treatment.

\textbf{Variations:}
\begin{itemize}\itemsep0pt
    \item \textit{Exhaustive}: Exhaustively describes all normal and abnormal findings.
    \item \textit{Minimal}: Selectively omits normal findings and focuses on a minimal list of key abnormal findings.
\end{itemize}

\textbf{Example 1:}
\begin{itemize}
    \item \textit{Exhaustive}: ``No significant technical limitations. Lung volumes are normal. No consolidative airspace disease. No pleural effusion or pneumothorax. No pulmonary nodule or mass noted. Pulmonary vasculature and the cardiomediastinal silhouette are within normal limits. Atherosclerosis in the thoracic aorta is noted.''
    \item \textit{Minimal}: ``No evidence of pneumothorax, pleural effusion, or pulmonary opacity. Cardiomediastinal silhouette is normal. Atherosclerosis in the thoracic aorta.''
\end{itemize}

\textbf{Example 2:}
\begin{itemize}
    \item \textit{Exhaustive}: ``No exam quality limitations. Patient positioning is appropriate. Trachea is midline. Image of the abdomen is within normal limits. No radiopaque lines, tubes, or implants. Hila are normal. Lung volumes are normal. Heart size is normal. Mediastinal contours are unremarkable. Lungs are clear. No effusion or pneumothorax. Bones are intact.''
    \item \textit{Minimal}: ``Heart size is normal. No evidence of pneumothorax, pleural effusion, or pulmonary opacity. No acute abnormality.''
\end{itemize}

\textbf{Example 3:}
\begin{itemize}
    \item \textit{Exhaustive}: ``QUALITY: Portable AP radiograph demonstrates low lung volumes with mild rotation. No motion artifact. The lungs are clear bilaterally. No pleural effusion or pneumothorax. The cardiomediastinal silhouette is within normal limits for this projection. Visualized osseous structures are unremarkable.''
    \item \textit{Minimal}: ``The lungs are clear bilaterally. No pleural effusion or pneumothorax. The cardiomediastinal silhouette is within normal limits. Visualized osseous structures are unremarkable.''
\end{itemize}

\textbf{Axis-Specific Constraints:} None.

\subsubsection{Granularity: Anatomical Granularity (Absolute)}
\textbf{Description:} Radiologists differ in the anatomical level at which they describe their findings. Some radiologists exhaustively enumerate findings for each anatomical substructure separately (e.g., ``pulmonary nodule in the right upper lobe anterior segment''), while others only enumerate high-level findings at the level of the organ or system (e.g., ``right lung pulmonary nodule'').

\textbf{Variations:}
\begin{itemize}\itemsep0pt
    \item \textit{High}: Describes findings for each anatomical substructure separately.
    \item \textit{Low}: Describes findings at the level of the organ or system.
\end{itemize}

\textbf{Example 1:}
\begin{itemize}
    \item \textit{High}: ``A pulmonary nodule is present in the right upper lobe anterior segment. A coronary artery stent is present in the left anterior descending artery. Small volume of fluid in the left posterior costophrenic sulcus. Right third through sixth healed rib fracture deformities. T9 anterior wedge compression fracture. Cholecystectomy clips are present.''
    \item \textit{Low}: ``Right lung pulmonary nodule. Prior coronary artery stenting. Small left pleural effusion. Healed rib fracture deformities. Thoracic compression fracture. Right upper quadrant surgical clips present.''
\end{itemize}

\textbf{Example 2:}
\begin{itemize}
    \item \textit{High}: ``The cardiac silhouette appears enlarged. Pulmonary vascular markings are prominent in the bilateral perihilar regions. There is mild interstitial prominence in the bilateral lower lungs. Trace fluid blunts the right posterior costophrenic sulcus. A left chest wall pacemaker is present with leads projecting over the cardiac silhouette. No focal airspace opacity. No pneumothorax.''
    \item \textit{Low}: ``Enlarged cardiac silhouette. Prominent pulmonary vasculature. Mild interstitial lung prominence. Trace right pleural effusion. Left-sided pacemaker present. No focal airspace opacity or pneumothorax.''
\end{itemize}

\textbf{Example 3:}
\begin{itemize}
    \item \textit{High}: ``The lungs are well expanded and clear without focal airspace opacity. The pulmonary vasculature is within normal limits. The costophrenic angles are sharp bilaterally. The diaphragms are smooth with normal contour. The cardiomediastinal silhouette is normal in size and shape. The trachea is midline. The hila are unremarkable. No pneumothorax is identified. The visualized osseous structures are intact.''
    \item \textit{Low}: ``Normal cardiomediastinal silhouette. Lungs are clear without consolidation, effusion, or pneumothorax. Pulmonary vasculature is within normal limits. No acute osseous abnormalities.''
\end{itemize}

\textbf{Axis-Specific Constraints:} None.

\subsubsection{Granularity: Quantitative Granularity (Absolute)}
\textbf{Description:} Radiologists differ in whether they include explicit numerical measurements (e.g., ``cardiothoracic ratio of 0.55'', ``2.4 cm at the apex'') or rely on qualitative descriptors (e.g., ``mild cardiomegaly'', ``moderate'').

\textbf{Variations:}
\begin{itemize}\itemsep0pt
    \item \textit{High}: Includes precise quantitative findings.
    \item \textit{Low}: Includes qualitative descriptions only.
\end{itemize}

\textbf{Example 1:}
\begin{itemize}
    \item \textit{High}: ``Left pleural effusion measuring 4 cm in height. Cardiothoracic ratio of 0.55.''
    \item \textit{Low}: ``Moderate left pleural effusion. Mild cardiomegaly.''
\end{itemize}

\textbf{Example 2:}
\begin{itemize}
    \item \textit{High}: ``There is an endotracheal tube whose distal tip is seen 6.2 cm above the carina appropriately sited.''
    \item \textit{Low}: ``There is an endotracheal tube whose distal tip is appropriately sited above the carina.''
\end{itemize}

\textbf{Example 3:}
\begin{itemize}
    \item \textit{High}: ``Moderate right apical pneumothorax has minimally decreased since yesterday. The maximum width at the apex measures 2.4 cm as compared to yesterday measuring 2.9 cm.''
    \item \textit{Low}: ``Moderate right apical pneumothorax has minimally decreased since yesterday.''
\end{itemize}

\textbf{Axis-Specific Constraints:} The \textit{Generator} should only modify the reference along this axis if the original reference contains precise quantitative findings, so as to avoid hallucinating new measurements when generating \textit{Quantitative Granularity (High)}. In practice, this constraint causes the \textit{Quantitative Granularity (High)} alternatives to be flagged as infeasible by the \textit{Verifier} for the vast majority of original references, and we report results only for \textit{Quantitative Granularity (Low)} in our experiments (Section~\ref{sec:experiments}).

\subsubsection{Language: Terminology (Relative)}
\textbf{Description:} Radiologists may use different terms to describe the same finding, depending on personal preference, training background, subspecialty, or institutional convention. These are near-synonymous phrasings that convey essentially the same clinical meaning.

\textbf{Variations:}
\begin{itemize}\itemsep0pt
    \item \textit{Original}: Retains the original terminology.
    \item \textit{Alternative}: Substitutes alternative terminology that conveys essentially the same clinical meaning.
\end{itemize}

\textbf{Example 1:}
\begin{itemize}
    \item \textit{Original}: ``No gross consolidation, atelectasis or infiltrate. No pleural fluid collection or pneumothorax. Cardiomediastinal silhouette is within normal limits.''
    \item \textit{Alternative}: ``The lungs are clear without evidence of focal airspace disease or consolidation. There is no evidence of pneumothorax or pleural effusion. The cardiac and mediastinal contours are within normal limits.''
\end{itemize}

\textbf{Example 2:}
\begin{itemize}
    \item \textit{Original}: ``No pulmonary opacity.''
    \item \textit{Alternative}: ``Lungs are clear.''
\end{itemize}

\textbf{Example 3:}
\begin{itemize}
    \item \textit{Original}: ``Cardiomegaly.''
    \item \textit{Alternative}: ``Enlarged cardiac silhouette.''
\end{itemize}

\textbf{Example 4:}
\begin{itemize}
    \item \textit{Original}: ``Bones are intact.''
    \item \textit{Alternative}: ``Osseous structures are within normal limits.''
\end{itemize}

\textbf{Example 5:}
\begin{itemize}
    \item \textit{Original}: ``No evidence of intraabdominal free air.''
    \item \textit{Alternative}: ``Visualized upper abdomen is normal.''
\end{itemize}

\textbf{Axis-Specific Constraints:} None.

\subsubsection{Language: Hedging (Relative)}
\textbf{Description:} Radiologists differ in how they express diagnostic uncertainty even when their underlying interpretations are similar. Different phrases can convey similar levels of uncertainty but use different linguistic constructions.

\textbf{Variations:}
\begin{itemize}\itemsep0pt
    \item \textit{Original}: Retains the original hedging language.
    \item \textit{Alternative}: Substitutes alternative hedging language that conveys essentially the same level of uncertainty.
\end{itemize}

\textbf{Example 1:}
\begin{itemize}
    \item \textit{Original}: ``Peripheral wedge-shaped opacity is worrisome for pulmonary embolism.''
    \item \textit{Alternative}: ``Peripheral wedge-shaped opacity is concerning for pulmonary embolism.''
\end{itemize}

\textbf{Example 2:}
\begin{itemize}
    \item \textit{Original}: ``Opacity consistent with atelectasis.''
    \item \textit{Alternative}: ``Opacity likely represents atelectasis.''
\end{itemize}

\textbf{Example 3:}
\begin{itemize}
    \item \textit{Original}: ``Right lung opacity, query pneumonia.''
    \item \textit{Alternative}: ``Right lung opacity may represent pneumonia.''
\end{itemize}

\textbf{Example 4:}
\begin{itemize}
    \item \textit{Original}: ``Left lung opacity, query pneumonia.''
    \item \textit{Alternative}: ``Left lung opacity suggestive of pneumonia.''
\end{itemize}

\textbf{Example 5:}
\begin{itemize}
    \item \textit{Original}: ``Bilateral perihilar opacities could represent pulmonary edema.''
    \item \textit{Alternative}: ``Bilateral perihilar opacities suggest possible pulmonary edema.''
\end{itemize}

\textbf{Axis-Specific Constraints:} The alternative reference must convey the same level of uncertainty as the original reference.

\subsubsection{Language: Negation (Absolute)}
\textbf{Description:} Radiologists differ in how they express negation, using either explicit statements about the absence of specific findings or implicit phrasings that convey the same information through positive statements about normality.

\textbf{Variations:}
\begin{itemize}\itemsep0pt
    \item \textit{Explicit}: Expresses negation explicitly (e.g., ``no pneumothorax, no pleural effusion'').
    \item \textit{Implicit}: Conveys negation implicitly through positive statements about normality (e.g., ``pleural spaces are clear'').
\end{itemize}

\textbf{Example 1:}
\begin{itemize}
    \item \textit{Explicit}: ``No pneumothorax. No pleural effusion.''
    \item \textit{Implicit}: ``Pleural spaces are clear.''
\end{itemize}

\textbf{Example 2:}
\begin{itemize}
    \item \textit{Explicit}: ``No focal consolidation. No pneumothorax. No pleural effusion.''
    \item \textit{Implicit}: ``The lungs are clear. Pleural spaces are normal.''
\end{itemize}

\textbf{Example 3:}
\begin{itemize}
    \item \textit{Explicit}: ``No mediastinal widening. No hilar adenopathy.''
    \item \textit{Implicit}: ``Mediastinal contours are within normal limits.''
\end{itemize}

\textbf{Axis-Specific Constraints:} \textit{Implicit} variations must clearly convey the same clinical information as the corresponding \textit{Explicit} negations. Ambiguous implicit statements that could be misinterpreted should be avoided.

\subsection{Prompts for the \textit{Generator} and the \textit{Verifier}}
\label{appendix:reref-prompts}

We instantiate \textsc{ReRef} with two prompts: a \textit{Generator} prompt that conditions an LLM on the original reference and the target axis-aligned variation, and a \textit{Verifier} prompt that conditions an LLM on the original reference, the target variation, and the candidate alternative reference produced by the \textit{Generator}, and asks it to assess (i)~whether the alternative reference preserves the clinical interpretation of the original (i.e., clinical equivalence) and (ii)~whether the perturbation is correctly simulated (i.e., the alternative actually reflects the target axis-aligned variation).
Both prompts are used with the system message ``You are a medical expert in radiology.''.
Below, fields enclosed in angle brackets (e.g., \texttt{<orig\_report>}, \texttt{<axis\_name>}) are placeholders that are populated at runtime with the original reference, the per-axis definitions, allowed variations, examples, and axis-specific constraints from Appendix~\ref{appendix:reref-axes}, and the target variation.

The \textit{Generator} prompt is constructed by concatenating the original reference, the per-axis description, allowed variations, and examples (Appendix~\ref{appendix:reref-axes}), the union of the default constraints listed below and the axis-specific constraint, and the target variation to generate.
The default constraints are:
\begin{itemize}\itemsep0pt
    \item Assume that all sections of the alternative reference other than the Findings and Impression (e.g., Indication, Technique) are identical to those of the original reference.
    \item Do \textbf{not} add any new information that is not substantiated by the original reference. In particular, if the original reference does not include the details required for perturbation along the specified axis, output the original reference unchanged.
    \item The alternative reference should remain clinically equivalent to the original reference.
    \item Introduce minimal changes beyond the specified axis along which the reference is being perturbed.
    \item Output only the alternative Findings and Impression sections, with no explanations or commentary.
\end{itemize}

\begin{custombox}[frametitle={Prompt for the \textit{Generator} LLM}]
\small
It is well-known that even for the same imaging study, radiologists can write the Findings and Impression sections in highly variable ways. Such variability can reflect differences in reporting style, structure, and interpretive emphasis, even when describing the same underlying findings. Given a ground-truth reference report for an imaging study, written by a radiologist, your task is to rewrite it along the specified axis of variation, while preserving all of the clinical content and interpretations present in the original report.

\textbf{========================~ORIGINAL REPORT~=========================}

\texttt{<orig\_report>}

\textbf{=========================~AXIS TO VARY~============================}

\begin{itemize}\itemsep0pt
    \item NAME: \texttt{<axis\_name>}
    \item DESCRIPTION: \texttt{<axis\_description>}
    \item VARIATIONS:
    \begin{itemize}\itemsep0pt
        \item \texttt{<variation\_1>}: \texttt{<variation\_1\_description>}
        \item \texttt{<variation\_2>}: \texttt{<variation\_2\_description>}
    \end{itemize}
    \item EXAMPLES: \texttt{<axis\_examples>} (JSON-formatted list of \{variation: example\} dictionaries from Appendix~\ref{appendix:reref-axes}).
\end{itemize}

\textbf{=====================~IMPORTANT CONSTRAINTS~======================}

\texttt{<default\_constraints + axis\_specific\_constraints>} (see itemized list above and Appendix~\ref{appendix:reref-axes}).

\textbf{=====================~VARIATION TO GENERATE~======================}

\texttt{<variation\_to\_generate>}
\end{custombox}

The \textit{Verifier} is provided with the original reference, the same axis-specific definitions, allowed variations, and examples that were shown to the \textit{Generator}, the target variation that the \textit{Generator} was asked to produce, and the alternative Findings and Impression sections that were sampled. It returns a structured response with four boolean fields plus a free-text \texttt{reasoning} field: \texttt{is\_clinically\_consistent} (whether the alternative reference is clinically equivalent to the original), \texttt{is\_valid\_perturbation} (whether the alternative reference correctly simulates the target axis-aligned variation), \texttt{does\_not\_contain\_required\_info} (whether the perturbation is infeasible because the original reference does not contain the information required to support it; e.g., the original lacks any quantitative measurements but the target variation is \textit{Quantitative Granularity (High)}), and \texttt{is\_already\_in\_expected\_form} (whether the original reference is already in the form expected after perturbation; e.g., the original is already structured but the target variation is \textit{Structure (Structured)}). When either of the latter two fields is \texttt{True}, the \textit{Verifier} returns \texttt{is\_valid\_perturbation = True}, so that the \textsc{ReRef} pipeline distinguishes infeasible cases from incorrectly perturbed ones; the \textit{Verifier}'s \texttt{reasoning} field allows us to subsequently flag and exclude infeasible studies from the corresponding axis-aligned analysis (Section~\ref{sec:experiments}).

\begin{custombox}[frametitle={Prompt for the \textit{Verifier} LLM}]
\small
You will be given a ground-truth reference report for an imaging study, written by a radiologist, and an alternative version of the report, generated by perturbing the ground-truth report along one or more axes of variation (e.g., report structure, terminology preferences, level of detail). Your task is to check if the alternative version is clinically consistent with the ground-truth report. Additionally, you will be given the axis along which the alternative version was meant to be perturbed, and you should check if the alternative version is a valid perturbation of the ground-truth report for the specified axis.

\textbf{IMPORTANT:}
\begin{itemize}\itemsep0pt
    \item The alternative version should \textbf{not} add any new information that is not substantiated by the original ground-truth reference report.
    \item Minor paraphrases and reordering (within and across sections) of clinical content are allowed, as long as they are clinically identical in meaning.
    \item There can be cases where the perturbation was not made, either because the ground-truth report (i)~does not contain sufficient information to make the perturbation without adding new information, or (ii)~contains sufficient information but is already in the form expected after the perturbation along the specified axis. In both cases, return \texttt{True} for the \texttt{is\_valid\_perturbation} field (otherwise, return \texttt{False}). For case~(i), also return \texttt{True} for \texttt{does\_not\_contain\_required\_info}; for case~(ii), also return \texttt{True} for \texttt{is\_already\_in\_expected\_form}. Provide a detailed explanation in the \texttt{reasoning} field.
    \item Examples of case~(i): the original report does not discuss technical quality, but the perturbation is to have it discussed; the original report does not contain any detailed quantitative measurements, but the perturbation is to include them.
    \item Examples of case~(ii): the original report is already structured, but the perturbation is to make it structured; the original report already exhaustively lists all normal and abnormal findings, but the perturbation is to make the list of findings exhaustive.
\end{itemize}

\textbf{========~GROUND-TRUTH REFERENCE REPORT~========}

\texttt{<orig\_report>}

\textbf{========~AXIS CHOSEN TO VARY~========}

\begin{itemize}\itemsep0pt
    \item NAME: \texttt{<axis\_name>}
    \item DESCRIPTION: \texttt{<axis\_description>}
    \item VARIATIONS:
    \begin{itemize}\itemsep0pt
        \item \texttt{<variation\_1>}: \texttt{<variation\_1\_description>}
        \item \texttt{<variation\_2>}: \texttt{<variation\_2\_description>}
    \end{itemize}
    \item EXAMPLES: \texttt{<axis\_examples>} (same JSON-formatted examples shown to the \textit{Generator}).
\end{itemize}

\textbf{========~VARIATION GENERATED~========}

\texttt{<variation\_generated>}

\textbf{========~ALTERNATIVE FINDINGS + IMPRESSION~========}

FINDINGS: \texttt{<alt\_findings>} \\
IMPRESSION: \texttt{<alt\_impression>}
\end{custombox}

\subsection{Inference Setup}
\label{appendix:reref-inference}

We instantiate both the \textit{Generator} and the \textit{Verifier} with o3 \citep{o3}, accessed via Microsoft Azure OpenAI under a setup compliant with the data use agreements for all of the datasets considered in our evaluations (Section~\ref{sec:experiments}).
The \textit{Generator} is sampled with temperature $0.7$ and a maximum output length of $16382$ tokens; the \textit{Verifier} is run at temperature $0$ to make its accept/reject decisions deterministic.
Outputs of both LLMs are parsed into Pydantic schemas via \texttt{instructor}, with the \textit{Generator} producing a \texttt{(findings, impression)} pair and the \textit{Verifier} producing the four boolean fields and \texttt{reasoning} field described in Appendix~\ref{appendix:reref-prompts}.
For each (study, axis-aligned variation) combination, the generation--verification loop runs for up to $5$ iterations; the loop terminates as soon as the \textit{Verifier} returns \texttt{is\_clinically\_consistent = True} and \texttt{is\_valid\_perturbation = True}, and otherwise the last sampled alternative reference is retained together with its \textit{Verifier} output, which we use downstream to filter out studies for which the \textit{Verifier} could not certify a clean axis-aligned alternative reference (Section~\ref{sec:experiments}).

\renewcommand\thetable{B\arabic{table}}
\renewcommand\thefigure{B\arabic{figure}}
\setcounter{table}{0}
\setcounter{figure}{0}
\section{Radiologist Annotation Guide for \textsc{ReRef}}
\label{appendix:annotation}
Here, we provide additional details on the radiologist annotation tasks outlined in Section~\ref{sec:annotation}. Below, we include the full annotation guide provided to each of the radiologist annotators, including the Overview, Task 1 (Realism) Annotation Guide, and Task 2 (Clinical Equivalence) Annotation Guide.

\begin{custombox}[frametitle={Overview of the Annotation Tasks}]
The quality of AI-generated radiology reports (in particular, the Findings and Impression sections) is often assessed using an automated evaluation metric, due to the challenges of obtaining clinician annotations at large scale. These automated metrics typically (i) compare each AI-generated report to a human-written reference report for the same imaging study and (ii) output a numerical score (e.g., between 0 and 1). For instance, an evaluation metric called GREEN (Ostmeier et al., 2024) is calculated as the proportion of clinical findings in the AI-generated report that match those in the human-written reference report, where the accuracy of each clinical finding is determined by an LLM (e.g., ChatGPT, Claude).

We are studying how robust such automated metrics are to stylistic variations in how different radiologists write radiology reports (e.g., report structure, expression of uncertainty, terminology preferences, level of detail). To do so, we made systematic edits to a large collection of human-written radiology reports from several chest X-ray (CXR) datasets (e.g., MIMIC-CXR, CheXpert Plus), simulating the various ways in which a practicing radiologist could have written the report. Each edit results in one alternative version of the original human-written CXR report. All such alternative reports are intended to be “clinically equivalent” to the original report that they were derived from, in terms of the key diagnostic findings and the downstream clinical decisions that would follow.

Your task is to evaluate whether these alternative reports are (i) realistic as standalone radiology reports (i.e., they look like what a practicing radiologist could have written) and (ii) clinically equivalent to the original reports. Your annotations will help us validate that the stylistic edits we study are grounded in real-world radiology reporting practice.

The annotation is organized into two phases, as described below.

\end{custombox}

\begin{custombox}[frametitle={Task 1 (Realism) Annotation Guide}]
\textbf{Goal}

Determine whether a given radiology report (the Findings and Impression sections) reads as a realistic, professionally written radiology report. Importantly, you should allow for the fact that certain stylistic and structural features of radiology reporting systematically vary across different practice environments.

\textbf{What You Will See}

Each annotation example will include a full radiology report, divided into two parts:
\begin{enumerate}
    \item Context: Indication, Comparison, Technique sections
    \item Report: Findings, Impression sections
\end{enumerate}

The realism rating should be based only on the ``Report'' portion. The ``Context'' portion is included solely as background information about the imaging study.

\textbf{What To Do}

For each report, provide the following information, allowing for the fact that certain stylistic and structural features of radiology reporting vary across different practice environments:
\begin{enumerate}
\item Realism Rating
    \subitem 3 — Realistic: Reads as a report written by a practicing radiologist in routine clinical care.
    \subitem 2 — Minor Issues: Mostly realistic, but contains subtle phrasing or formatting choices that are slightly unusual (though clinically correct).
    \subitem 1 — Major Issues: Contains major errors, incoherent phrasing, or unnatural elements that a practicing radiologist would not write in a standard radiology report.
\item Free-Text Comments (required for 1 and 2): Please briefly explain your reasoning.
\end{enumerate}

\textbf{Important Notes (Please Read)}
\begin{itemize}
\item ``Realistic” does not mean “you would personally write the report this way“. Radiologists have diverse reporting styles across training backgrounds, practice environments, etc. The question is whether some practicing radiologist might plausibly have written such a report.
\item As noted above, the realism rating should only be given based on the Findings and Impression sections of the radiology report, and not the other auxiliary sections (Comparison, Indication, Technique).
\item Because all reports have undergone a de-identification process (often automated in an imprecise manner), the reports shown to you may contain redactions (e.g., ``XXXX'' or ``\_\_\_\_'') that render the reports awkward or grammatically wrong. Please ignore such de-identification artifacts when determining the realism of each report.
\end{itemize}

\end{custombox}

\begin{custombox}[frametitle={Task 2 (Clinical Equivalence) Annotation Guide}]
\textbf{Goal}

Given a pair of radiology reports about the same imaging study, determine whether they present the same clinical assessment---i.e., whether a treating physician reading either report would reach the same diagnostic conclusions and clinical decisions.

\textbf{What You Will See}

Each annotation example will include the following parts, all pertaining to the same CXR study:
\begin{enumerate}
    \item Context: Indication, Comparison, Technique sections
    \item Report A: Findings, Impression
    \item Report B: Findings, Impression
\end{enumerate}

The equivalence rating should be based only on the ``Report A'' and ``Report B'' portions. The two reports may differ in style, structure, level of detail, or terminology. The ``Context'' portion is included solely as background information about the imaging study and is assumed to stay the same between the two reports.

\textbf{What To Do}

For each pair of reports, provide the following information:
\begin{enumerate}
    \item Clinical Equivalence Rating
        \subitem \textbf{Equivalent}: A treating physician reading either report would identify the same findings, reach the same diagnostic conclusions, and make the same clinical decisions. Differences between the reports are limited to style, structure, terminology, or level of descriptive detail, and do not affect what clinical actions would follow. Examples of differences that are still ``equivalent'':
        \begin{itemize}
            \item One report lists both normal and abnormal findings exhaustively, while the other mentions only the key abnormalities.
            \item One report specifies ``airspace opacity'', while the other says ``consolidation''.
            \item One report includes precise measurements, while the other uses qualitative descriptions (e.g., ``moderate'').
        \end{itemize}
        \subitem \textbf{Minor Discrepancy}: The reports agree on the main diagnoses and no critical findings are missing from either report, but there is a small difference that a treating physician might notice. The difference is unlikely to change clinical management. Examples:
        \begin{itemize}
            \item One report mentions a minor incidental finding (e.g., mild degenerative changes) that the other omits.
            \item One report characterizes a finding as ``mild'', while the other says ``mild-to-moderate''.
            \item One report includes a differential possibility that the other does not, but the primary assessment is the same.
        \end{itemize}
        \subitem \textbf{Major Discrepancy}: The reports differ in major ways that could lead a treating physician to different clinical decisions. Examples:
        \begin{itemize}
            \item One report identifies a finding that the other does not mention at all (e.g., a pulmonary nodule, pneumothorax, or consolidation).
            \item The reports differ in the characterization of a finding in a way that would change urgency or management (e.g., ``small'' vs.\ ``large'' effusion).
            \item One report suggests a diagnosis that the other contradicts or omits entirely.
        \end{itemize}
    \item Free-Text Comments (required for Minor and Major Discrepancy): Describe the specific differences that led you to your rating.
\end{enumerate}

\textbf{Important Notes (Please Read)}
\begin{itemize}
    \item Two reports can differ substantially in wording, structure, and level of detail while still being clinically equivalent. The question is not whether the reports are identical, but whether they would lead to the same clinical actions.
    \item Please do not determine the equivalence rating based on the realism of each report. Determining the realism of each report is reserved as a separate annotation task. \\
    \textit{Example}: Suppose that two reports contain exactly the same text for describing the findings, but one report writes all of them under the Findings section (i.e., no Impression section), while the other writes all of them under the Impression section (i.e., no Findings section). The former reads less realistic than the latter, as radiology reports generally always have an Impression section, but since they contain identical findings and would lead to the same actions, this pair should be labeled as ``Equivalent''.
    \item A report that mentions fewer findings is not automatically discrepant. Some radiologists routinely omit normal findings or reduce descriptive detail without altering the clinical interpretation and conclusion. Only flag omissions that could cause a treating physician to miss something actionable.
    \item Differences in how uncertainty is expressed (e.g., ``concerning for'' vs.\ ``suspicious for'') are equivalent unless they convey meaningfully different levels of confidence that would change clinical decision making.
    \item Because all reports have undergone a de-identification process (often automated in an imprecise manner), the reports shown to you may contain redactions (e.g., ``XXXX'' or ``\_\_\_\_'') that render the reports awkward or grammatically wrong. Please ignore such de-identification artifacts when determining the equivalence between reports in each pair.
\end{itemize}

\end{custombox}

\renewcommand\thetable{C\arabic{table}}
\renewcommand\thefigure{C\arabic{figure}}
\setcounter{table}{0}
\setcounter{figure}{0}
\section{Additional Details on Datasets}
\label{appendix:datasets}
\begin{table}[t]
\centering
\caption{Summary of the four CXR datasets used in our evaluation (Section~\ref{sec:experiments}).
For each dataset, we report the final number of examples retained
(i.e., those with a Findings section in the reference report) and summary statistics
on the number of images per study.}
\label{tab:dataset-summary}
\begin{tabular}{lrr*{4}{r}}
\toprule
& & & \multicolumn{4}{c}{\# Images in Study} \\
\cmidrule(lr){4-7}
Dataset & \# Studies & Total Images & Mean $\pm$ Std & Median & Min & Max \\
\midrule
CheXpert Plus & $61$       & $72$       & $1.18 \pm 0.46$ & $1$ & $1$ & $3$  \\
IU X-ray       & $590$      & $1{,}223$  & $2.07 \pm 0.28$ & $2$ & $2$ & $4$  \\
MIMIC-CXR     & $2{,}882$  & $4{,}633$  & $1.61 \pm 0.70$ & $1$ & $1$ & $5$  \\
ReXGradient-160K   & $10{,}000$ & $17{,}029$ & $1.70 \pm 0.73$ & $2$ & $1$ & $29$ \\
\midrule
Total         & $13{,}533$ & $22{,}957$ & ---             & --- & --- & ---  \\
\bottomrule
\end{tabular}
\end{table}

In Table~\ref{tab:dataset-summary}, we provide a summary of the number of imaging studies included in each of the four datasets used in our evaluation (Section~\ref{sec:experiments}). 
For all datasets, we pre-extract the key sections from each reference report (i.e., the Indication, Comparison, Technique, Findings, and Impression sections) to identify the Findings section to use for RRG model evaluation. 
For all datasets except MIMIC-CXR, all of the sections are already available in parsed form, so we use them directly as available. 
For MIMIC-CXR, we extract all sections by prompting Claude Sonnet 4.5~\citep{anthropic2025claude-sonnet-4-5} using the prompt below, which is adapted from the prompt used by \citet{llava-rad}\footnote{See Supplementary Table 6 in \href{https://arxiv.org/pdf/2403.08002}{arxiv.org/pdf/2403.08002}}:
\begin{custombox}[frametitle={Prompt for Section Extraction from MIMIC-CXR}]
\small
\textbf{System Prompt:}

You are an expert medical assistant AI capable of modifying clinical documents to user specifications. You make minimal changes to the original document to satisfy user requests. You never add information that is not already directly stated in the original document.

\textbf{Section Extraction Prompt:}

Extract the following sections from the input radiology report: [{section\_names}]. Leave an extracted section as "N/A" if it does not exist in the original report. The output should be in JSON format. An Indication section can refer to the History, Indication or Reason for Study sections in the original report. An Impression section can include the Recommendations and Conclusion sections in the original report. The extracted sections should be verbatim from the original report and should not overlap with each other. Remove any linebreaks and whitespace that have been added only for better readability.

Examples of inputs and expected outputs (when e.g., extracting the [`Examination', `Indication', `Findings', `Impression'] sections):

\textbf{INPUT:}\\
EXAMINATION: XR CHEST AP PORTABLE\\
INDICATION: Small right apical pneumothorax after lung biopsy.\\
FINDINGS: Single portable view of the chest was obtained. Copared with 10:42 AM. The small right apical pneumothorax has decreased slightly in size, the improvement best appreciated laterally where it now measures 10 mm compared to 14 mm before. At the lung apex it now measures 1.6 and compared to 2.1 cm previously. A subtle right apical pulmonary contusion is grossly stable. Minor chest wall emphysema along the right exilla has not changed significant delay. There is no metastatic shift. No pleural effusion is evident.

\textbf{OUTPUT:}
\{\\
      "EXAMINATION": "XR CHEST AP PORTABLE.",\\
      "INDICATION": "Small right apical pneumothorax after lung biopsy.",\\
      "FINDINGS": "Single portable view of the chest was obtained. The small right apical pneumothorax measures 10mm. At the lung apex it measures 1.6cm. A subtle right apical pulmonary contusion is grossly stable. Minor chest wall emphysema is noted along the right exilla. There is no metastatic shift. No pleural effusion is evident.",\\
      "IMPRESSION": "N/A"\\
\}

**Here is the report that you should process**:\\
\{Report\}
\end{custombox}

\renewcommand\thetable{D\arabic{table}}
\renewcommand\thefigure{D\arabic{figure}}
\setcounter{table}{0}
\setcounter{figure}{0}
\section{Additional Details on Models}
\label{appendix:models}
Here, we provide additional details on how we generate the reports from each model considered in our study. 
By default, we generate the reports from each model via greedy decoding.
For the models that support multi-image inputs, we provide as many CXR views as feasible from each imaging study, up to 10 images (or whatever number of images each model can support). 
If more images are present, we prioritize the inclusion of frontal over lateral images.
We note that there are no imaging studies that have more than 10 images in the test sets for MIMIC-CXR, CheXpert Plus, and IU X-ray. 
In ReXGradient-160K, there are only 2 out of the 10K imaging studies in the test set that contain more than 10 images for an imaging study (one with 21 images, the other with 29 images).
For all inference runs, we use up to 8 NVIDIA A6000 GPUs, each with 48GB of memory. 

\subsection{MAIRA-2 \citep{maira-2}}
\label{appendix:maira-2}

For MAIRA-2 \citep{maira-2}, we use the official checkpoint on HuggingFace\footnote{\href{https://huggingface.co/microsoft/maira-2}{huggingface.co/microsoft/maira-2}} and follow the instructions in the model card for generating the Findings section in the ``ungrounded'' setting (i.e., the model does not generate bounding boxes as evidence for visual findings localizable in the CXR images). 
For Findings generation, we use the prompts provided in Table B.1 of the paper, providing the Indication, Technique, and Comparison sections of the report as additional context when available for the given study.
Any missing sections are marked as ``N/A''.
As MAIRA-2 supports inputs with up to one frontal image and one lateral image (i.e., maximum of two images corresponding to different views), we always include both views if available. 
If one of the two views is missing, we provide as input whichever view is available.

\subsection{MedGemma-4B \citep{medgemma}}
\label{appendix:medgemma}

For MedGemma-4B \citep{medgemma}, we use the official checkpoint for the instruction-tuned model on HuggingFace\footnote{\href{https://huggingface.co/google/medgemma-4b-it}{huggingface.co/google/medgemma-4b-it}}. 
For Findings generation, we use our own custom prompt, 
additionally providing the Indication, Technique, and Comparison sections if available for the given study: ``Generate the FINDINGS section of the radiology report for this CXR image.\textbackslash n\textbackslash nBelow are the prefilled sections of the report that you can consider:\textbackslash n\textbackslash nINDICATION: $\langle$indication$\rangle$\textbackslash n\textbackslash nTECHNIQUE: $\langle$technique$\rangle$\textbackslash n\textbackslash nCOMPARISON: $\langle$comparison$\rangle$\textbackslash n\textbackslash nDO NOT generate anything other than the FINDINGS section of the report, and DO NOT repeat the details of the prefilled sections in any way.''
As MedGemma-4B is originally developed for single-image input settings, we only provide the frontal view of the CXR as input. 

\subsection{CheXOne \citep{chexone}}
\label{appendix:chexone}

For CheXOne \citep{chexone}, we use the official checkpoint on HuggingFace\footnote{\href{https://huggingface.co/StanfordAIMI/CheXOne}{huggingface.co/StanfordAIMI/CheXOne}}.
For Findings generation, we use the prompts provided in Figure S12 of the paper: ``Given the indication $\langle$indication$\rangle$, write an example findings section for the CXR.''
We always sample generations from the model in its ``reasoning mode'' (see Section 2.7 of the paper), where we append the following at the end of the main prompt: ``Please reason step by step and put your final answer within \texttt{\textbackslash boxed\{\}}''. 
For imaging studies with multiple CXR views, we select up to 10 images to provide as input, prioritizing the frontal images over lateral images. 

\subsection{MedVersa \citep{medversa}}
\label{appendix:medversa}

For MedVersa \citep{medversa}, we use the official checkpoint on HuggingFace\footnote{\href{https://huggingface.co/hyzhou/MedVersa}{huggingface.co/hyzhou/MedVersa}}. 
We select one of the 10 prompts for Findings generation listed in Table 5 of their arXiv preprint version\footnote{\href{https://arxiv.org/abs/2405.07988}{arxiv.org/abs/2405.07988}} of the paper: ``How would you characterize the findings from $\langle$<img0>...<imgN>$\rangle$?''
For imaging studies with multiple CXR views, we select up to 10 images to provide as input, prioritizing the frontal images over lateral images. 

\subsection{CheXagent-2-3B \citep{chexagent}}
\label{appendix:chexagent}

For CheXagent-2-3B \citep{chexagent}, we use the official checkpoint on HuggingFace\footnote{\href{https://huggingface.co/StanfordAIMI/CheXagent-2-3b}{huggingface.co/StanfordAIMI/CheXagent-2-3b}}.
For Findings generation, we use the prompt provided in the demo script in the official GitHub repository\footnote{\href{https://github.com/Stanford-AIMI/CheXagent/blob/main/model\_chexagent/chexagent.py\#L63}{github.com/Stanford-AIMI/CheXagent/blob/main/model\_chexagent/chexagent.py}}: ``Given the indication $\langle$indication$\rangle$, write a structured findings section for the CXR.''
For multi-image inputs, we include only up to two images per study, following the authors' approach on their CheXbench evaluation suite \citep{chexagent}\footnote{\href{https://github.com/Stanford-AIMI/CheXagent/blob/main/evaluation\_chexbench/axis\_3\_text\_generation/run\_findings\_generation.py\#L35}{github.com/Stanford-AIMI/CheXagent/blob/main/evaluation\_chexbench/axis\_3\_text\_generation/run\_findings\_generation.py}}.

\subsection{Med-CXRGen \citep{med-cxrgen}}
\label{appendix:med-cxrgen}

For Med-CXRGen \citep{med-cxrgen}, we use the official checkpoints on HuggingFace, which are released as two separate models for the Findings (Med-CXRGen-F\footnote{\href{https://huggingface.co/X-iZhang/Med-CXRGen-F}{huggingface.co/X-iZhang/Med-CXRGen-F}}) and Impression (Med-CXRGen-I\footnote{\href{https://huggingface.co/X-iZhang/Med-CXRGen-I}{huggingface.co/X-iZhang/Med-CXRGen-I}}) sections.
For Findings generation, we use the prompt provided in the official codebase: ``Provide a detailed description of the findings in the radiology image.'' For Impression generation, we use the analogous prompt with ``findings'' replaced by ``impression''.
Following the reference implementation, we additionally append the Indication, Technique, and Comparison sections of the report as clinical context to the prompt when available.
For imaging studies with multiple CXR views, we follow the authors' multi-image strategy and stitch up to 4 images horizontally with a 10-pixel black separator between adjacent images, prioritizing the key (frontal) view first followed by remaining frontal views over lateral views.

\subsection{Libra \citep{libra}}
\label{appendix:libra}

For Libra \citep{libra}, we use the official checkpoint on HuggingFace\footnote{\href{https://huggingface.co/X-iZhang/libra-v1.0-7b}{huggingface.co/X-iZhang/libra-v1.0-7b}}.
For Findings generation, we use the prompt provided in the official codebase: ``Provide a detailed description of the findings in the radiology image.''
As with Med-CXRGen, we additionally append the Indication, Technique, and Comparison sections as clinical context to the prompt when available.
Libra is designed to take as input a paired set of CXR images consisting of one image from the current study and one image from a prior study; as our test sets do not include prior studies, we follow the authors' recommended fallback and duplicate the current image as a placeholder for the prior.
Accordingly, we provide as input only the key (frontal) view of the current study, since Libra was not trained or evaluated on multi-view input settings.

\subsection{LLaVA-Rad \citep{llava-rad}}
\label{appendix:llava-rad}

For LLaVA-Rad \citep{llava-rad}, we use the official checkpoint on HuggingFace\footnote{\href{https://huggingface.co/microsoft/llava-rad}{huggingface.co/microsoft/llava-rad}}, loaded on top of the \texttt{lmsys/vicuna-7b-v1.5} base language model as specified by the authors.
For Findings generation, we use the prompt provided in the official demo script: ``Describe the findings of the chest x-ray.''
As LLaVA-Rad was developed and evaluated only on single-image inputs, we provide as input only the key (frontal) view of the CXR.

\subsection{CheX-MIMIC \citep{chexpertplus}}
\label{appendix:chex-mimic}

For CheX-MIMIC \citep{chexpertplus}, we use the official Findings generation baseline checkpoint released alongside CheXpert Plus \citep{chexpertplus} on HuggingFace\footnote{\href{https://huggingface.co/IAMJB/chexpert-mimic-cxr-findings-baseline}{huggingface.co/IAMJB/chexpert-mimic-cxr-findings-baseline}}. 
As CheX-MIMIC is a purely image-conditioned encoder-decoder model that does not accept any text prompt, no instruction or clinical context is provided alongside the input image.
As CheX-MIMIC was developed and evaluated only on single-image inputs, we provide as input only the key (frontal) view of the CXR.

\renewcommand\thetable{E\arabic{table}}
\renewcommand\thefigure{E\arabic{figure}}
\setcounter{table}{0}
\setcounter{figure}{0}
\section{Additional Results: \textit{Reporting Practice Variations Can Impact Conclusions from Model Comparison} (Section~\ref{sec:results-nonuniform-impact})}
\label{appendix:ranking-reversals}
\begin{figure*}[t!]
    \centering
    \begin{tabular}{@{}c@{}c@{}c@{}c@{}}
        \includegraphics[width=\linewidth]{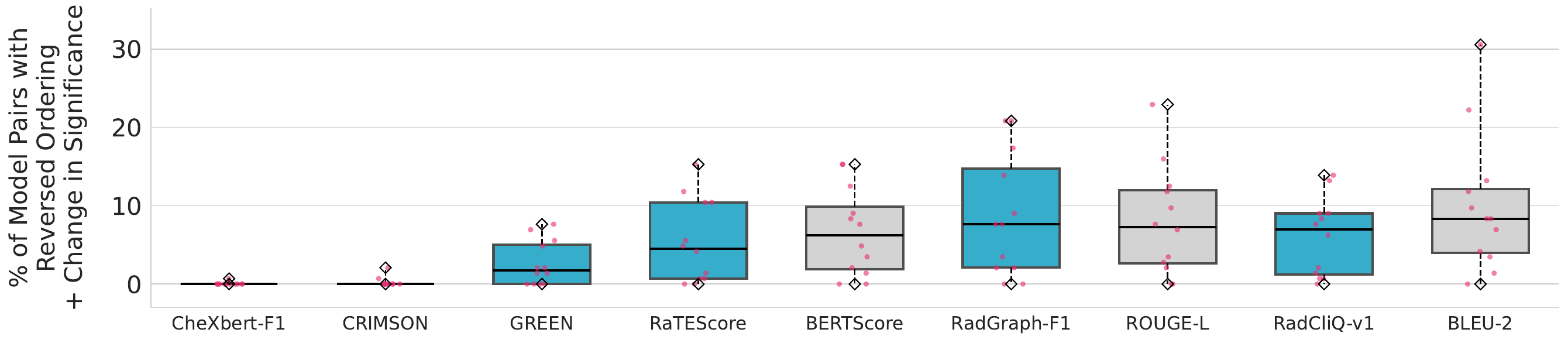}
    \end{tabular}
    \vspace{-7pt}
    \caption{Model rankings can be sensitive to the reporting practices of the reference (Section~\ref{sec:results-nonuniform-impact}), even when we limit to reversals that also change the 95\% bootstrap CI-based comparison for each model pair. 
    Aggregating the results across all datasets and model pairs, we find that for 7 out of 9 metrics, pairwise ranking reversals are often accompanied by a change in the 95\% bootstrap CI-based comparison for each pair. 
    Each box plot shows, across the 12 axis-aligned perturbations, the percentage of all $4 \times \binom{9}{2} = 144$
    (dataset, model pair) comparisons whose pairwise ordering by mean score reverses relative to the original reference \textit{and} whose 95\% bootstrap CI-based comparison also changes (Section~\ref{sec:results-nonuniform-impact}). 
    The diamond-shaped markers indicate the max and min values across all perturbations. 
    The cyan-colored box plots indicate the clinical metrics.
    }
    \label{fig:ranking-sensitivity-significant}
\end{figure*}

Figure~\ref{fig:ranking-sensitivity}(b) includes a pairwise ranking reversal whenever the ordering of two models by average score reverses relative to the original reference, irrespective of the magnitude of the underlying score gap. 
As some of these gaps are small enough to fall within noise, we additionally report in Figure~\ref{fig:ranking-sensitivity-significant} the subset of reversals for which the 95\% bootstrap CI-based comparison for each pair also changes, i.e., where the perturbation alters the conclusion one would draw about the pair rather than only its ordering. 
When pooled across all metrics, datasets, and perturbations, roughly 60\% of all pairwise ranking reversals (in Figure~\ref{fig:ranking-sensitivity}(b)) are accompanied by a change in the CI-based comparison, and the metrics most affected remain broadly similar (median reversal in Figure~\ref{fig:ranking-sensitivity-significant}: RadGraph-F1: 7.6\%; RadCliQ-v1: 7.0\%; RaTEScore: 4.7\%). 
The two notable exceptions are CRIMSON and CheXbert-F1, for which the reversals are rarely accompanied by a significant change in pairwise relative mean scores.
For CRIMSON, such a result is consistent with its comparatively low overall sensitivity to reporting practice variations, as observed in Figure~\ref{fig:global-metric-sensitivity}.

\renewcommand\thetable{F\arabic{table}}
\renewcommand\thefigure{F\arabic{figure}}
\setcounter{table}{0}
\setcounter{figure}{0}
\section{Additional Results: Compound Reporting Practice Variations}
\label{appendix:compound-perturbations}
\begin{figure*}[t!]
    \centering
    \begin{tabular}{@{}c@{}}
         \begin{subfigure}{0.98\linewidth}
            \includegraphics[width=\linewidth]{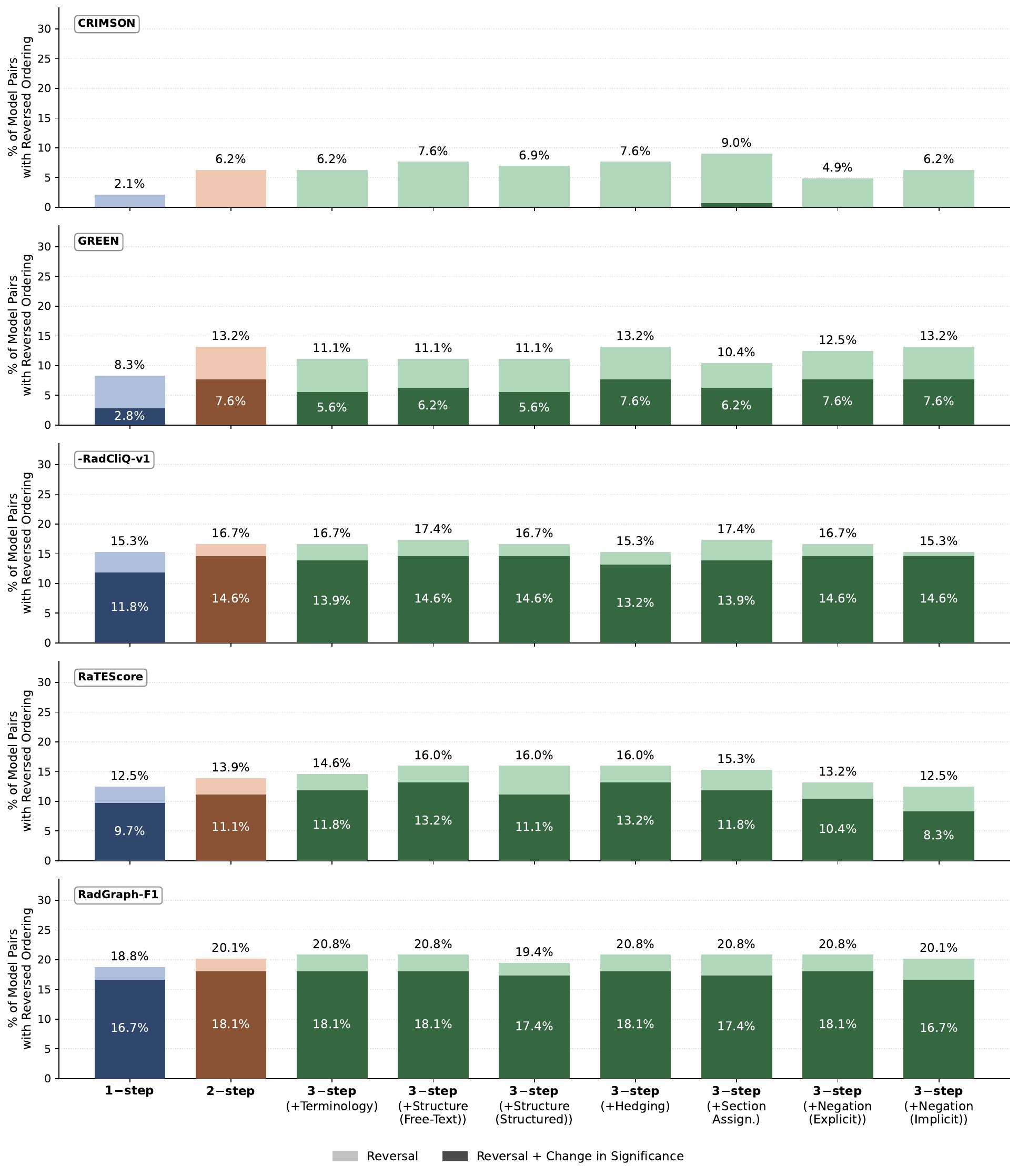}
         \end{subfigure}
    \end{tabular}
    \caption{Compound reporting practice perturbations along multiple axes (2-step, 3-step) generally result in larger deviations from the model rankings under the original reference, compared to the single-axis perturbation (1-step) setting. 
    Here, we show the results under the \textit{minimalist} persona, which composes \textit{Findings Scope (Minimal)} with \textit{Anatomical Granularity (Low)}, followed in the 3-step setting by one additional axis (indicated on the $x$-axis; Appendix~\ref{appendix:compound-perturbations}). 
    Each bar reports the percentage of all $4 \times \binom{9}{2} = 144$ (dataset, model pair) comparisons whose pairwise ordering reverses relative to the original reference. 
    The darker lower segment is the subset of those reversals for which the gap between the two models' mean scores also changes between the three cases distinguished by a 95\% bootstrap confidence interval in relative score: significantly favoring one model, significantly favoring the other, or statistically indistinguishable from zero (tie). 
    }
    \label{fig:ranking-sensitivity-compound-minimalist}
\end{figure*}

\begin{figure*}[t!]
    \centering
    \begin{tabular}{@{}c@{}}
         \begin{subfigure}{0.98\linewidth}
            \includegraphics[width=\linewidth]{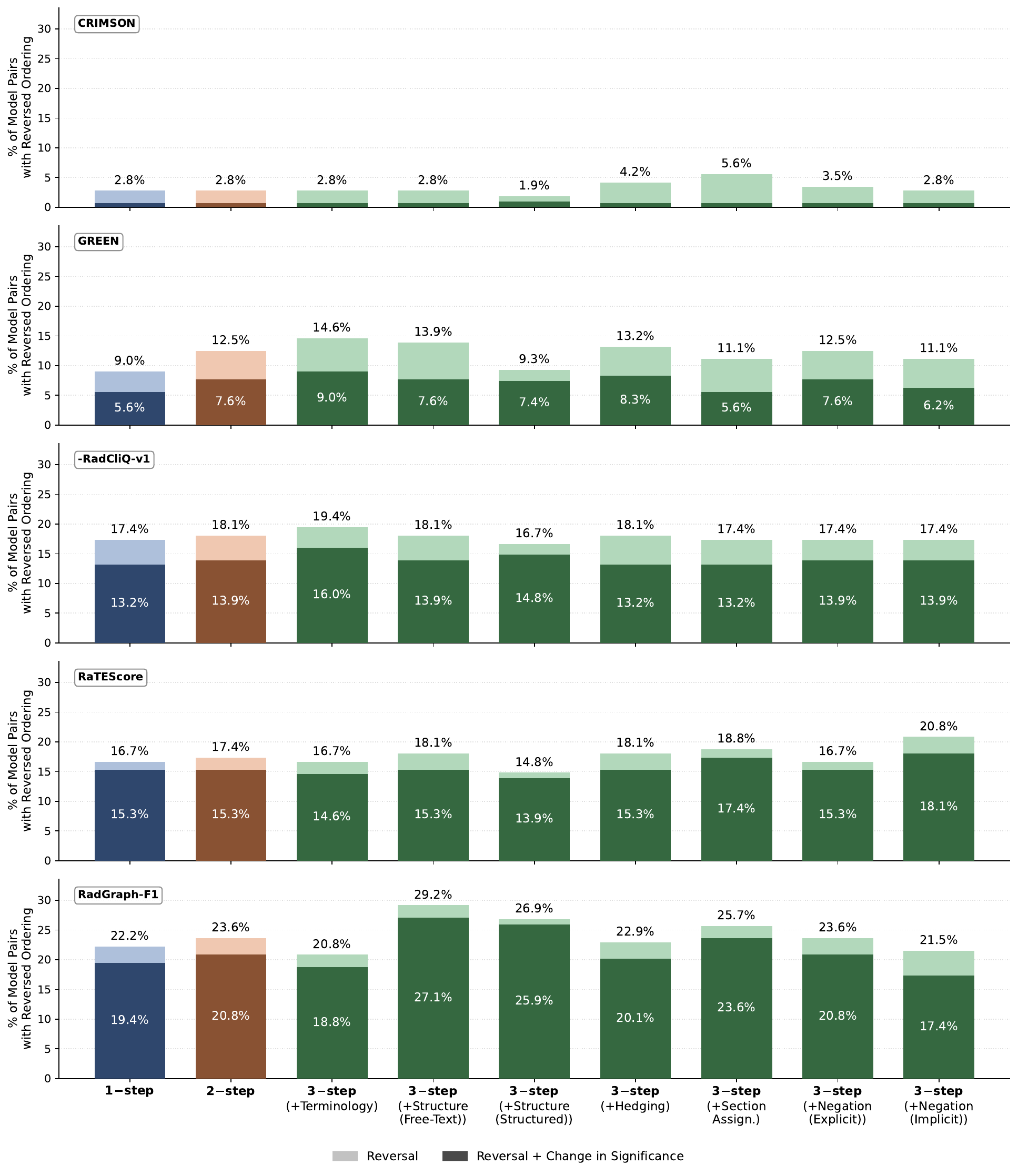}
         \end{subfigure}
    \end{tabular}
    \caption{Compound reporting practice perturbations along multiple axes (2-step, 3-step) generally result in larger deviations from the model rankings under the original reference, compared to the single-axis perturbation (1-step) setting. 
    Here, we show the results under the \textit{maximalist} persona, which composes \textit{Findings Scope (Exhaustive)} with \textit{Anatomical Granularity (High)}, followed in the 3-step setting by one additional axis (indicated on the $x$-axis; Appendix~\ref{appendix:compound-perturbations}). 
    Each bar reports the percentage of all $4 \times \binom{9}{2} = 144$ (dataset, model pair) comparisons whose pairwise ordering reverses relative to the original reference. 
    The darker lower segment is the subset of those reversals for which the gap between the two models' mean scores also changes between the three cases distinguished by a 95\% bootstrap confidence interval in relative score: significantly favoring one model, significantly favoring the other, or statistically indistinguishable from zero (tie). 
    }
    \label{fig:ranking-sensitivity-compound-maximalist}
\end{figure*}

While we primarily measure metric sensitivity to reporting practice variations along a \textit{single} axis in our main experiments (Section~\ref{sec:results}), real-world radiology reports likely reflect \textit{multiple} reporting practices simultaneously and in a \textit{correlated} manner. 
For instance, a radiologist who prefers to write concise reports that only discuss the key abnormal findings (\textit{Findings Scope (Minimal)}) may also be less detailed in describing their findings across anatomical substructures (\textit{Anatomical Granularity} (Low)). 
In contrast, a radiologist who prefers to write exhaustive reports that thoroughly discuss all normal and abnormal findings (\textit{Findings Scope (Exhaustive)}) may also be more thorough in describing their findings across all anatomical substructures (\textit{Anatomical Granularity (High)}). 

Based on this intuition, we repeat the experiments in Section~\ref{sec:results} under the setup where references are subject to \textit{compound} reporting practice variations (i.e., applying multiple axis-aligned perturbations). 
To do so, we use the above example and consider two radiologist \textit{personas}: (i) a \textit{minimalist} who aligns with \textit{Findings Scope (Minimal)} and \textit{Anatomical Granularity (Low)}, and (ii) a \textit{maximalist} who aligns with \textit{Findings Scope (Exhaustive)} and \textit{Anatomical Granularity (High)}. 
For each persona, we also allow for variations along the remaining axes (e.g., \textit{Structure}, \textit{Terminology}, \textit{Hedging}).
Thus, the alternative references we use for analysis are subject to \textit{up to three} reporting practice perturbations. 

Given that the specific order in which the \textit{Findings Scope} and \textit{Anatomical Granularity} perturbations are applied matters, we account for both directions when computing the scores under the compound alternatives. 
For instance, in the \textit{minimalist} setting, we perturb each reference report once in the order of
\textit{Findings Scope (Minimal)} $\rightarrow$ \textit{Anatomical Granularity (Low)}, and once in the opposite direction, and then average the scores calculated against each resulting alternative reference. 
For the 3-step setting, we apply one further axis-aligned perturbation (e.g., \textit{Terminology}) to each of these 2-step alternatives, again averaging over the two orderings of the first two axes. 

Since not every perturbation is feasible for every imaging study (as determined by the \textit{Verifier} in \textsc{ReRef}), we score the model-generated reports on each study against the most heavily perturbed alternative reference available for it within a given setting.
In the 3-step setting, we score against the 3-step alternative where feasible, followed by the 2-step alternatives, then the single-axis (\textit{Findings Scope} and \textit{Anatomical Granularity}) alternatives (averaged when both are feasible), and finally their original-reference scores, as considered in Section~\ref{sec:results-nonuniform-impact}.
In the 2-step setting, we apply an analogous procedure, starting from the 2-step alternatives. 

Compared to the single-axis perturbation case, we observe that deviations in model ranking from the original-reference setting tend to be greater as we perturb the references further (1-step $\rightarrow$ 2-step $\rightarrow$ 3-step), while the exact trends vary across metrics 
(Figures~\ref{fig:ranking-sensitivity-compound-minimalist}--\ref{fig:ranking-sensitivity-compound-maximalist}). 
Notably, some 3-step perturbations lower the reversal rate relative to the 2-step case, likely because the third perturbation partially counteracts the effect of the first two perturbations. 
For instance, \textit{Negation (Explicit)} can \textit{expand} the discussion of findings in the reference by adding explicit statements about absent abnormalities (e.g., ``pleura appear normal'' $\rightarrow$ ``no pneumothorax, no pleural effusion''), which can counteract the first two perturbations under the \textit{minimalist} persona and lowers the pairwise ranking reversal rate (e.g., CRIMSON: 6.2\% (2-step) $\rightarrow$ 4.9\% (3-step; \textit{Negation (Explicit)}); Figure~\ref{fig:ranking-sensitivity-compound-minimalist}).



\end{document}